\documentclass[acmlarge]{acmart}
\AtBeginDocument{%
 }

\usepackage{hyperref}   % ← 先に読む
\usepackage{hyperxmp}   % ← 後で読む

\usepackage{subcaption}
\usepackage{adjustbox}

\makeatletter
\newcommand{\figcaption}[1]{\def\@captype{figure}\caption{#1}}
\newcommand{\tblcaption}[1]{\def\@captype{table}\caption{#1}}
\makeatother

\setcopyright{none}
\begin{document}

\title{C-AAE: Compressively Anonymizing Autoencoders for Privacy-Preserving Activity Recognition in Healthcare Sensor Streams}

 % to Counteract Digital Temptations and Deceptive Tactics

% laptop

\author{Ryusei Fujimoto}
% \orcid{0009-0001-7993-0117}
\affiliation{%
  \institution{Kyushu University}
  \city{Fukuoka}
  \country{Japan}
}

\author{Yugo Nakamura}
\orcid{0000-0002-8834-5323}
\authornote{Corresponding author}
\email{y-nakamura@ait.kyushu-u.ac.jp}
\affiliation{%
  \institution{Kyushu University}
  \city{Fukuoka}
  \country{Japan}
}

\author{Yutaka Arakawa}
\orcid{0000-0002-7156-9160}
\affiliation{%
  \institution{Kyushu University}
  \city{Fukuoka}
  \country{Japan}
}

% \renewcommand{\shortauthors}{Y.Nakamura, et al.}



\begin{abstract}
Wearable accelerometers and gyroscopes encode fine-grained behavioural signatures that can be exploited to re-identify users, making privacy protection essential for healthcare applications. We introduce C-AAE, a compressive anonymizing autoencoder that marries an Anonymizing AutoEncoder (AAE) with Adaptive Differential Pulse-Code Modulation (ADPCM). The AAE first projects raw sensor windows into a latent space that retains activity-relevant features while suppressing identity cues. ADPCM then differentially encodes this latent stream, further masking residual identity information and shrinking the bitrate.
Experiments on the MotionSense and PAMAP2 datasets show that C-AAE cuts user re-identification F1 scores by 10-15 percentage points relative to AAE alone, while keeping activity-recognition F1 within 5 percentage points of the unprotected baseline. ADPCM also reduces data volume by roughly 75 \%, easing transmission and storage overheads. These results demonstrate that C-AAE offers a practical route to balancing privacy and utility in continuous, sensor-based activity recognition for healthcare.

\end{abstract}

% %\begin{figure}[t]
% \centering
% \includegraphics[width=0.99\linewidth]{image/hbs-overview_aiot-sys0.pdf}
% \vspace{-3mm}
% \caption{
% Concept of Health Behavioral Security:
% (a) Odysseus resists the Sirens with the help of his crew.
% (b) In modern society, cognitive vulnerabilities expose individuals to digital temptations.
% (c) AIoT as System 0 senses, filters, and nudges in real time, building health resilience that mitigates harmful influences and encourages healthier behaviors.
% }
% \label{fig:overview}
% %\vspace{-3mm}
% %\end{figure}


\maketitle

\vspace{-1mm}

% %1章

\section{Introduction}

Continuous sensing with wearables and smartphone‐embedded inertial sensors underpins a growing range of mobile health and remote–patient-monitoring applications~\cite{bib:wearable_har,bib:wearable_uid}. 
Because these streams encode fine-grained behavioural signatures, they raise pressing privacy concerns in clinical and everyday-care settings. Effective privacy protection must therefore balance two competing requirements: (i)~suppress user-identifiable cues and (ii)~preserve activity information needed for downstream health analytics. Prior work tackles this trade-off through differential privacy~\cite{bib:dp} or representation-learning approaches such as Replacement AutoEncoder (RAE) and Anonymizing AutoEncoder (AAE)~\cite{bib:AAE,bib:MLP-AAE}. However, strong differential-privacy budgets often cripple recognition accuracy~\cite{bib:fujimoto}, and RAE necessitates costly manual rules to decide which behaviours to mask.

AAE can, in principle, anonymise sensor data to random-guessing levels, yet its protection degrades when behavioural diversity grows; earlier evaluations covered only four activities and relied chiefly on convolutional classifiers~\cite{bib:cnn-1,bib:cnn-2}. We find that anonymisation weakens further when those CNNs are replaced by widely used tabular classifiers such as LightGBM~\cite{bib:lgbm} or Random Forest~\cite{bib:rf}.

To overcome these limitations, we present \textbf{Compressive-AAE (C-AAE)}, a two-stage framework that first applies AAE and then compresses the latent sequence with Adaptive Differential Pulse-Code Modulation (ADPCM). ADPCM encodes the difference between successive samples while adapting the quantisation rate, thereby shrinking data volume and masking residual identity cues. Earlier work applied the simpler DPCM variant in a differential-privacy setting~\cite{bib:adpcm}; we adopt ADPCM because its adaptive step minimises reconstruction error—an attractive property for health-care analytics conducted at the edge.

When ADPCM operates on AAE‐processed streams, recognition accuracy decreases only marginally, yet user re-identification accuracy drops sharply—even as the number of activity classes or the choice of classifier grows. The resulting bitstream is also \emph{approximately 75 \%} smaller than the original, easing on-device storage and back-haul traffic—practical benefits for resource-constrained health IoT deployments.

We evaluate C-AAE on the MotionSense~\cite{bib:motion-sense} and PAMAP2~\cite{bib:pamap2} datasets. Relative to a strong AAE baseline, C-AAE lowers re-identification F1 by roughly 10–15 percentage points while keeping activity-recognition F1 within 5 points of the unprotected baseline. All processing can be executed locally on embedded hardware, further aligning with privacy regulations in clinical practice.

Our contributions are three-fold:
\begin{itemize}
 \item We introduce \emph{Compressive-AAE}, a privacy-preserving pipeline that couples AAE with ADPCM to anonymise inertial data while retaining clinical utility.
 \item We show that ADPCM substantially reduces residual identity information—especially under diverse activity sets and non-CNN classifiers—without materially harming recognition accuracy.
 \item Extensive experiments on two public benchmarks confirm a \textasciitilde75 \% reduction in data volume and a 10–15 \% drop in re-identification accuracy, suggesting that C-AAE is well-suited for edge-level deployment in real-world mobile health systems.
\end{itemize}

\section{Preliminaries}
\label{sec:prelim}

\subsection{Differential Privacy}
Traditional privacy-preserving methods such as user-ID removal and $k$-anonymity only guard against specific attacks~\cite{bib:id_removal,bib:k-anonymity}. 
Differential Privacy (DP) overcomes this limitation by providing rigorous, quantitative guarantees that do not depend on the attacker’s background knowledge~\cite{bib:any-prevent}.

\begin{definition}[$\epsilon$-Differential Privacy]
\label{def:dp}
Let $\mathcal{D}$ be the domain of databases and $D_1,D_2\!\in\!\mathcal{D}$ two adjacent databases differing in at most one record. 
A randomized mechanism $\mathcal{M}\!:\!\mathcal{D}\!\to\!\mathcal{R}$ is $\epsilon$-DP if for every measurable set $\mathcal{S}\!\subseteq\!\mathcal{R}$,
\[
 \Pr\!\bigl[\mathcal{M}(D_1)\!\in\!\mathcal{S}\bigr]
 \;\le\; e^{\epsilon}\,
 \Pr\!\bigl[\mathcal{M}(D_2)\!\in\!\mathcal{S}\bigr].
\]
\end{definition}

Smaller $\epsilon$ implies stronger privacy. 
A widely used instantiation is the \textit{Laplace mechanism}, which adds noise drawn from a Laplace distribution:
\[
 \mathrm{Lap}(x;1/\epsilon)=\frac{\epsilon}{2}\exp\!\bigl(-\epsilon\lvert x\rvert\bigr).
\]

Because of its simplicity and formal guarantees, the Laplace mechanism is commonly adopted in sensor-based activity-recognition pipelines~\cite{bib:trade-off-1,bib:har,bib:har-2,stirapongsasuti2024preserving}.

\subsection{Signal Compression Techniques}
Pulse Code Modulation (PCM) digitises analog signals via uniform sampling. 
Differential PCM (DPCM) improves compression by encoding sample‐to-sample differences, exploiting temporal redundancy. 
Adaptive DPCM (ADPCM) further adjusts the quantisation step size online, making it well-suited to real-time streams from resource-constrained devices and balancing fidelity against bitrate.

\subsection{Generative Representation Learning}
Generative Adversarial Networks (GANs): A generator and discriminator are trained adversarially; the generator synthesises data while the discriminator distinguishes real from fake, yielding high-fidelity samples~\cite{bib:gan,bib:gan2}.

Autoencoders (AEs): An encoder–decoder pair reconstructs inputs through a bottleneck, learning compact latent codes~\cite{bib:autoencoder}. 
The \emph{Anonymising AutoEncoder (AAE)} augments this with discriminators that minimise mutual information between latent codes and user identities, thus preserving task utility while suppressing identity cues~\cite{bib:AAE}.

% ===========================
% 3. RELATED WORK
% ===========================
\section{Related Work}
\label{sec:related}

\subsection{Differential Privacy for Activity Recognition}
Differential privacy (DP) has been extensively explored for anonymizing sensor data to protect user identities. Traditional approaches such as user ID removal and $k$-anonymity safeguard only against specific known attacks~\cite{bib:id_removal,bib:k-anonymity}. By contrast, DP provides rigorous, quantifiable guarantees that are resilient to arbitrary attacks~\cite{bib:any-prevent}. DP has been applied in eye-tracking with VR headsets~\cite{bib:eye_track}, wearable devices, and smart-home environments~\cite{bib:trade-off-1,bib:har,bib:har-2,stirapongsasuti2024preserving}. A persistent challenge is balancing privacy protection and data utility. A weighted-noise DP mechanism reduced user re-identification risk but did not reach practical activity-recognition accuracy~\cite{bib:fujimoto}.

\subsection{Signal Compression-based Anonymization}
Signal-processing techniques such as Pulse Code Modulation (PCM) and Differential PCM (DPCM) offer another path to anonymizing continuous sensor streams. Parra-Arnau \textit{et al.} adapted DPCM for data anonymization, reducing distortion by 30\,\% compared with traditional methods while satisfying DP constraints~\cite{bib:adpcm}. Fixed quantization, however, limits adaptability. We therefore adopt Adaptive DPCM (ADPCM), which dynamically adjusts quantization levels, further optimizing the privacy–utility trade-off.

\subsection{Generative Approaches for Privacy Preservation}
Generative Adversarial Networks (GANs) generate synthetic data that preserve statistical properties while masking sensitive attributes. Menasria \textit{et al.} proposed Private GANs (PGAN1, PGAN2) that achieve high activity-recognition accuracy (90--100\,\%) and very low user-recognition rates (3--4\,\%)~\cite{bib:pgan}. Hybrid GAN+DP models further enhance privacy~\cite{bib:gan,bib:gan2,bib:hybrid-dp-gan}, but require large training datasets and user-specific pre-training, complicating deployment.

\subsection{Autoencoder-based Anonymization Techniques}
The Anonymizing AutoEncoder (AAE) anonymizes sensor data on-device while retaining activity-recognition accuracy~\cite{bib:AAE}. Malekzadeh \textit{et al.} reported 92.9\,\% activity-recognition accuracy with only 1.8\,\% user-recognition accuracy. Lightweight variants such as MLP-AAE on ESP32 show feasibility for resource-constrained devices~\cite{bib:MLP-AAE}. Nevertheless, performance degrades with larger activity sets (e.g.\ PAMAP2) or non-CNN classifiers (Random Forest, LightGBM), revealing scalability limitations.

\subsection{Research Gap and Motivation}
DP methods provide strong guarantees but often sacrifice utility. DPCM/ADPCM offer efficient compression yet remain limited by fixed or heuristic quantizers. GAN-based approaches need substantial training data, and AAE struggles to generalize across diverse datasets and classifiers. To bridge these gaps, we propose \emph{Compressive-AAE (C-AAE)}, which jointly optimizes ADPCM quantization and AAE representation learning to deliver privacy-preserving activity recognition with high utility across heterogeneous conditions.

\section{Compressive-AAE}
\label{sec:c-aae}

This section presents the proposed Compressive-AAE (C-AAE) framework, which integrates two complementary techniques—Anonymizing AutoEncoder (AAE) and Adaptive Differential Pulse Code Modulation (ADPCM)—to achieve privacy-preserving activity recognition. We first describe the system model and threat model that motivate the need for privacy-aware data processing. Then, we introduce the application criteria that define the balance between privacy and utility. Following this, we provide an overview of the C-AAE framework and explain how each component functions, including a detailed explanation of ADPCM. Finally, we describe the system implementation of the C-AAE-based application, covering its end-to-end data flow from sensing to feedback.

\subsection{System Model}
\begin{figure}[t]
  \centering
  \includegraphics[clip, width=0.5\columnwidth]{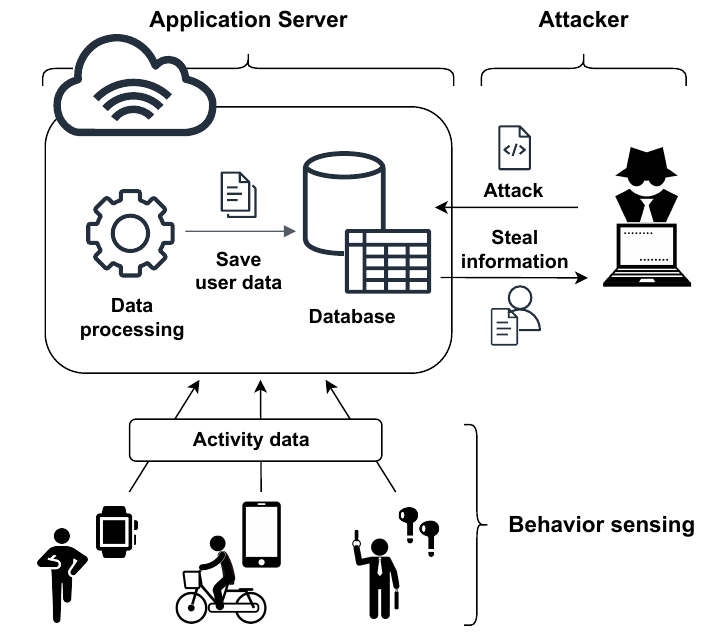}
  \vspace{1mm}
  \caption{Envisioned System Model}
  \vspace{-3mm} % Adjust spacing between the figure and text
  \label{fig:model}
\end{figure}

The envisioned system model is illustrated in Figure~\ref{fig:model}. We consider a real-time activity tracking application that collects user behavior through wearable and mobile devices equipped with sensors such as accelerometers and gyroscopes. The collected activity data is preprocessed on the user's edge device to reduce privacy risks before being transmitted to the application server.

On the server side, the received data undergoes further processing to estimate the user's activity, and both the activity labels and processed data are stored in a database. Since this data may contain sensitive personal information, it must be protected from external threats. In particular, we assume the presence of attackers who may attempt to intercept transmissions or access the stored data in order to infer private information.

To address these risks, our approach focuses on preprocessing sensor data in a way that minimizes user-identifiable information while retaining the information necessary for activity recognition. The system model shown in Figure~\ref{fig:model} serves as the basis for the design and implementation of the proposed Compressive-AAE framework.

% 図1

\subsection{Threat Model}
As illustrated in Figure~\ref{fig:model}, we assume an attacker who attempts to infer private information about a target user by analyzing the sensor data collected from wearable or mobile devices. Specifically, the attacker aims to identify the user (user re-identification) or extract sensitive attributes, such as behavioral patterns or demographics, from the transmitted sensor data.

We consider the following capabilities for the attacker:
\begin{itemize}
\item The attacker can intercept the transmitted sensor data between the user's edge device and the server, or access data stored on the server.
\item The attacker may possess background knowledge, such as prior data samples from known users or statistical information about the user population (e.g., activity distributions).
\item The attacker can apply machine learning techniques, including supervised classification models, to perform user identification or behavior inference.
\end{itemize}

To mitigate the risks posed by such attackers, this study focuses on protecting the \emph{user identity} as private information. We assume that the application scenario allows the user's \emph{activity labels} to be utilized for service provision, whereas the user ID should remain undisclosed. In other words, we aim to preserve privacy by ensuring that, even if the activity can be inferred, it is not possible to determine which user performed it.

This threat model reflects realistic privacy concerns in activity recognition applications, where sensor data may unintentionally reveal personal identities or sensitive behavioral traits. Our proposed framework, C-AAE, is designed to minimize the information relevant to user identity while retaining information necessary for activity classification.

\subsection{Application Criteria}

In this study, the condition for ensuring user privacy and eliminating privacy risks is defined as follows: ``a user can only be identified with a probability equivalent to random chance''~\cite{bib:eye_track}. Specifically, if the application has $n$ users, the probability that an attacker can identify a user from activity data, regardless of the estimator used, must not exceed $\frac{1}{n}$.

To summarize, the activity recognition framework used in this system model (application) must satisfy the following two requirements:

\begin{enumerate}
  \item The activity recognition accuracy must be maintained after data processing, within a margin of 5\% compared to unprocessed data.
  \item Users can only be identified with a probability equivalent to random chance.
\end{enumerate}

In Section~\ref{sec:evaluation}, the results of conventional methods and the proposed method are evaluated based on these two requirements. This study collectively defines these two requirements as the ``Application Criteria.''

\subsection{Overview of C-AAE}\label{subsec:caae_overview}

Compressive-AAE (C-AAE) is a two-stage, privacy-preserving preprocessing framework for wearable-sensor streams. 

\paragraph{Stage 1: Anonymising AutoEncoder (AAE).} 
A raw window \(x\in\mathbb{R}^{6\times128}\) is mapped to a latent tensor
\(z\in\mathbb{R}^{d\times16}\).
The encoder maximises mutual information with the target activity \(y\) while
minimising mutual information with the user identity \(u\).
Experiments show that AAE alone anonymises well, but residual
user-specific cues remain when behaviours are diverse.

\paragraph{Stage 2: Adaptive Differential Pulse-Code Modulation (ADPCM).} 
ADPCM, a parameter-free yet signal-adaptive encoder, processes the latent stream by
quantising temporal differences
\(\Delta z_t = z_t - \hat z_{t-1}\)
with step size
\(s_t = f\!\bigl(\lVert\Delta z_{t-1}\rVert_2\bigr)\).
This second step (a) compresses data by \(\approx 4\times\) and 
(b) disrupts any residual identity traces left by AAE.

Combining the learned transformation (AAE) with the adaptive,
non-learned encoding (ADPCM) yields stronger anonymisation than either
component alone, while keeping activity-recognition accuracy within the
5\,\% margin required by the Application Criteria
(Section~\ref{sec:evaluation}).

\subsection{Adaptive Differential Pulse-Code Modulation (ADPCM)}%
\label{subsec:adpcm_detail}

Differential PCM (DPCM) encodes each sample as the difference from a
predictor, but a \emph{fixed} quantiser step causes precision loss in
high-variance segments and redundancy in flat segments.
\textbf{ADPCM} updates the step size online, e.g.\
\[
 s_t = \alpha\,s_{t-1} + (1-\alpha)\,\beta\,\lvert\Delta z_{t-1}\rvert,
 \quad 0 < \alpha < 1,
\]
providing fine quantisation for low-variance periods and coarse
quantisation for bursts~\cite{bib:adpcm-1,bib:adpcm-2}.

Our experiments show that applying ADPCM to the AAE latent tensor
\(z\) (i) preserves salient activity features, (ii) removes
high-frequency user idiosyncrasies, and (iii) yields a compact code
suitable for low-bandwidth transmission.

Signal-domain transforms such as ADPCM also help satisfy
\(\epsilon\)-differential-privacy guarantees by bounding the sensitivity
of the encoded stream~\cite{bib:adpcm}. Thus, integrating ADPCM into
C-AAE meets our Application Criteria: activity-recognition F1 within
5\,\% of baseline and user re-identification no better than random
guessing.

\subsection{System Implementation}
\label{subsec:system}

% 図2
\begin{figure}[t]
  \centering
  \includegraphics[clip, width=0.5\columnwidth]{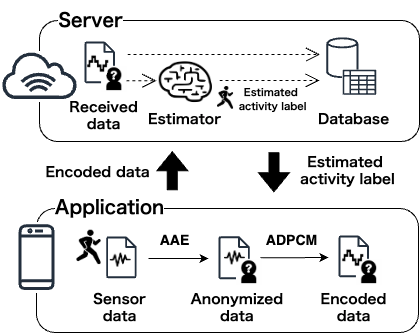}
  \vspace{1mm}
  \caption{System Implementation Overview}
  \vspace{-3mm} % Adjust spacing between the figure and text
  \label{fig:system}
\end{figure}

Figure~\ref{fig:system} shows the implementation structure of the activity management application based on the C-AAE framework, as envisioned in the system model in Figure~\ref{fig:model}. The system has two main components: an edge device and a cloud-based server.

On the edge device, the system continuously collects sensor data using built-in accelerometers and gyroscopes. This raw sensor data is first processed by an Anonymizing AutoEncoder (AAE), which converts the input into a latent representation that suppresses user-identifiable features while retaining activity-relevant information.

The anonymized output is then encoded using Adaptive Differential Pulse Code Modulation (ADPCM), which compresses the data and further obfuscates identifiable characteristics. The encoded data is transmitted to the server via a network connection.

On the server side, the received ADPCM-encoded data is decoded and passed to an activity recognition estimator (e.g., a classifier model), which predicts the user's activity label. The predicted label and associated encoded data are stored in a database. The activity label is also returned to the edge device and presented to the user through the application interface.

This implementation pipeline enables real-time, privacy-preserving activity tracking while reducing communication overhead and preserving user anonymity. It is designed for lightweight execution on edge devices and scalable processing on cloud servers.

\begin{table*}[t]
\renewcommand{\arraystretch}{1.0}
 \caption{Overview of the Datasets Used}
 \label{table:dataset}
 \centering
 \begin{adjustbox}{max width=0.7\textwidth} 
 \begin{tabular}{cccc}
  \hline
  Dataset & Sensor Location & Activities (Labels) & Number of Subjects \\
  \hline \hline
  MotionSense & Pants pocket & Walking (wlk) & 24 \\
        & & Ascending stairs (ups) & \\
        & & Descending stairs (dws) & \\
        & & Jogging (jog) & \\
        & & Standing (std) & \\
        & & Sitting (sit) & \\
  \hline
  PAMAP2   & Hand & Lying & 9 \\
       & Chest & Sitting & \\
       & Ankle & Standing & \\
       & & Walking & \\
       & & Running & \\
       & & Cycling & \\
       & & Nordic walking & \\
       & & Ascending stairs & \\
       & & Descending stairs & \\
       & & Vacuum cleaning & \\
       & & Ironing & \\
       & & Rope jumping & \\
  \hline
 \end{tabular}
  \end{adjustbox}
\end{table*}

% Table 5
\begin{table*}[t]
\renewcommand{\arraystretch}{1.3}
 \caption{Features Used as Input to Machine Learning Algorithms}
 \label{table:features}
 \centering
  \begin{adjustbox}{max width=\textwidth} 
 \begin{tabular}{cc}
  \hline
  Feature & Description \\
  \hline \hline
  Mean & Mean value within the window data \\
  Standard Deviation (SD) & Value representing the variance within the window data \\
  Max & Maximum value within the window data \\
  Min & Minimum value within the window data \\
  Dominant Frequency & Frequency with the largest amplitude spectrum in the frequency domain of the window data \\
  \hline
 \end{tabular}
  \end{adjustbox}
\end{table*}

\section{Evaluation}
\label{sec:evaluation}

This section outlines the experimental setup and evaluation methods used to assess the effectiveness of the proposed method, C-AAE, in comparison with the Anonymizing AutoEncoder (AAE)~\cite{bib:AAE}. To highlight the limitations of noise-based anonymization, we first present a preliminary evaluation using a conventional differential privacy technique. We then describe the datasets, estimators, and training procedures used in the comparative experiments, followed by the evaluation results and analysis.

\subsection{Datasets}

This study uses two publicly available datasets, MotionSense and PAMAP2, for comparative experiments. Details are summarized in Table~\ref{table:dataset}.

MotionSense contains data collected from the 3-axis accelerometer and 3-axis gyroscope sensors of an iPhone 6s. The data were recorded at 50$\mathrm{Hz}$ while the smartphone was placed in the user's front pants pocket. It includes six activity types performed by 24 subjects aged 18 to 46. For this study, the data were segmented into 2.56-second windows for training and testing.

PAMAP2 includes recordings of 18 activities performed by 9 subjects aged 23 to 32, using inertial measurement units (IMUs) worn on the hand, chest, and ankle, along with a heart rate monitor. We used only the 3-axis accelerometer and gyroscope data from the IMUs, selecting 12 activity types for the experiments. The data were sampled at 100$\mathrm{Hz}$ and segmented into 1.28-second windows.

\subsection{Experiment Details}
\subsubsection{Estimator and Training Method}

The comparative experiments in this study were conducted using the datasets shown in Table~\ref{table:dataset}. For MotionSense, as used in prior research~\cite{bib:AAE}, both activity and user recognition employed CNN as the estimator. The input data consisted of the vector magnitudes of the 3-axis accelerometer data and the 3-axis gyroscope data. The primary difference from the prior research is that the training data for constructing the AAE and the training data for the estimator also include static activity data in this study.

Next, for PAMAP2, experiments were conducted using CNN~\cite{bib:cnn-1}\cite{bib:cnn-2} as the estimator for both activity and user recognition, as well as using machine learning algorithms (LightGBM (LGBM)~\cite{bib:lgbm}, Random Forest (RF)~\cite{bib:rf}, and Logistic Regression (LR)~\cite{bib:lr}). In the experiments where CNN was used as the estimator, tests were conducted both when a single IMU from the three locations was used and when multiple IMUs were combined. The input data consisted of the vector magnitudes of the 3-axis accelerometer data and the 3-axis gyroscope data for each IMU. In the experiments where machine learning algorithms were used as estimators, only cases where multiple IMUs were combined were tested. Here, the input data format differed between the AAE and the estimators. For the AAE, the input data consisted of all the 3-axis accelerometer and 3-axis gyroscope data from the selected IMUs, which were anonymized. For the estimators, features were extracted for each window as shown in Table~\ref{table:features} and used as input.

For the application of ADPCM, it was performed immediately after anonymization by AAE in all experiments. In the case of machine learning algorithms, feature extraction was conducted after the application of ADPCM.

\subsubsection{Technical Validation of ADPCM}
\label{subsec:adpcm}

The results of the technical validation of ADPCM for activity data are shown in Table~\ref{table:adpcm}. For this validation, the PAMAP2 dataset, as described in Table~\ref{table:dataset}, was used as the input data, and data from all three IMUs were utilized. The estimators used were LGBM, RF, and LR. The results show that for LGBM and RF, the activity recognition mean F1 was almost maintained, while the user recognition mean F1 was reduced by approximately 12$\sim$16\%. For LR, although the activity recognition mean F1 decreased by 10\%, the user recognition mean F1 was reduced by approximately 30\%. Although the application criteria have not yet been fully achieved, these results indicate that ADPCM is effective for privacy preservation while maintaining the utility of activity data.

% Table 6
\begin{table}[t]
\renewcommand{\arraystretch}{1.3}
 \caption{Technical Validation Results of ADPCM}
 \label{table:adpcm}
 \centering
 \begin{tabular}{lccc}
  \hline
   & LGBM & RF & LR \\
  \hline \hline
  (Before) Activity mean F1 & 0.914 & 0.916 & 0.889 \\
  (After) Activity mean F1 & \textbf{0.879} & \textbf{0.873} & \textbf{0.776} \\
  \hline
  (Before) User mean F1 & 0.717 & 0.689 & 0.558 \\
  (After) User mean F1 & \textbf{0.555} & \textbf{0.572} & \textbf{0.254} \\
  \hline
 \end{tabular}
\end{table}

\subsubsection{Evaluation Method}

The evaluation methods differed depending on whether CNN or machine learning algorithms were used as the estimators. First, when CNN was used as the estimator, for each experiment, both the case where AAE was applied for estimation and the case where C-AAE was applied for estimation were evaluated. For each case, the training and testing were conducted five times, and the mean F1 score of the five tests was used as the evaluation metric. Next, when machine learning algorithms were used as estimators, Stratified 10-fold cross-validation~\cite{bib:skcv} was employed for model evaluation. The F1 score was used as the evaluation metric for these models.

The configurations differed for activity recognition and user recognition for the training, validation, and test data. Two participants' data were randomly selected for activity recognition as test data, while the remaining data were used as training and validation data. For user recognition, the data were randomly split such that all users were included in both the training/validation and test datasets at a ratio of 4:1.

\subsection{Preliminary Evaluation of Differential Privacy Techniques}
\label{subsec:dp}

Before presenting the evaluation results of the proposed method (C-AAE), we conducted a baseline experiment using a conventional differential privacy technique—the Laplace mechanism—to assess the inherent trade-off between privacy and utility. The purpose of this evaluation is to illustrate the limitations of noise-based anonymization methods in the context of activity recognition.

Table~\ref{table:dp-motionsense} shows the results for MotionSense when noise based on the Laplace distribution was added to the input features. As shown in Table~\ref{table:dataset}, the dataset includes data from 24 subjects. Thus, Requirement 2 of the Application Criteria is satisfied only when user recognition accuracy is below approximately 4.17\%. The results indicate that this condition is met only when $\epsilon = 0.1$, but at the cost of significant performance degradation—activity recognition accuracy drops to around 23\%, violating Requirement 1.

Similarly, Table~\ref{table:dp-pamap2} shows the results for PAMAP2, which includes 9 subjects (threshold for Requirement 2: 11.1\%). Again, user recognition accuracy meets the criterion only when $\epsilon = 0.1$, but activity recognition accuracy decreases sharply to the range of 13.6–32.9\%. These results highlight the difficulty of satisfying both privacy and utility requirements simultaneously when using traditional noise-based privacy-preserving mechanisms.

Overall, this preliminary evaluation confirms that conventional differential privacy methods entail a strong trade-off between activity recognition performance and privacy protection, which the proposed method (C-AAE) seeks to overcome.

% Table 3
\begin{table*}[t]
\renewcommand{\arraystretch}{1.0}
 \caption{Results of Applying Differential Privacy (MotionSense). SD = Standard Deviation.}

 \label{table:dp-motionsense}
 \centering
 \begin{tabular}{cccccccccc}
  \hline
  Target & $\epsilon = $ 0.1 & 0.3 & 0.5 & 1.0 & 1.5 & 2.0 & 2.5 & 3.0 & No Noise \\
  \hline \hline
  Activity mean F1 & 0.230 & 0.492 & 0.596 & 0.755 & 0.796 & 0.827 & 0.845 & 0.851 & \textbf{0.903} \\
 SD& 0.0237 & 0.0113 & 0.0144 & 0.00694 & 0.00777 & 0.00708 & 0.00695 & 0.0113 & 0.0180 \\
  \hline
  User mean F1 & \textbf{0.0350} & 0.0785 & 0.166 & 0.336 & 0.425 & 0.462 & 0.483 & 0.493 & 0.532 \\
 SD& 0.00215 & 0.00381 & 0.00482 & 0.00709 & 0.00442 & 0.00831 & 0.00543 & 0.00626 & 0.0188 \\
  \hline
 \end{tabular}
\end{table*}

% Table 4
\begin{table}[t]

\renewcommand{\arraystretch}{1.0}
 \caption{Results of Applying Differential Privacy (PAMAP2). SD = Standard Deviation.}
 \label{table:dp-pamap2}
 \centering
 \begin{adjustbox}{max width=\textwidth} 
 \begin{tabular}{ccccccccccc}
  \hline
  Sensor Location & Target & $\epsilon = $ 0.1 & 0.3 & 0.5 & 1.0 & 1.5 & 2.0 & 2.5 & 3.0 & No Noise \\
  \hline \hline
  Hand & Activity mean F1 & 0.139 & 0.309 & 0.389 & 0.497 & 0.571 & 0.599 & 0.613 & 0.640 & \textbf{0.712} \\
    &SD& 0.0275 & 0.0230 & 0.00780 & 0.00613 & 0.0126 & 0.0185 & 0.0315 & 0.0174 & 0.00765 \\
    & User mean F1 & \textbf{0.101} & 0.186 & 0.275 & 0.397 & 0.552 & 0.596 & 0.649 & 0.703 & 0.914 \\
    &SD& 0.00886 & 0.00745 & 0.0104 & 0.0220 & 0.00357 & 0.00650 & 0.0150 & 0.0104 & 0.0106 \\
  \hline
  Chest & Activity mean F1 & 0.173 & 0.371 & 0.493 & 0.552 & 0.628 & 0.642 & 0.655 & 0.647 & \textbf{0.662} \\
     &SD& 0.00679 & 0.0151 & 0.00362 & 0.0124 & 0.00725 & 0.0116 & 0.0112 & 0.0211 & 0.0257 \\
    & User mean F1 & \textbf{0.109} & 0.180 & 0.268 & 0.431 & 0.534 & 0.607 & 0.660 & 0.712 & 0.918 \\
    &SD& 0.00622 & 0.00941 & 0.0198 & 0.0107 & 0.00979 & 0.00172 & 0.00372 & 0.00900 & 0.00936 \\
  \hline
  Ankle & Activity mean F1 & 0.136 & 0.407 & 0.499 & 0.609 & 0.624 & 0.651 & 0.655 & 0.670 & \textbf{0.676} \\
     &SD& 0.0110 & 0.00495 & 0.0113 & 0.0153 & 0.0209 & 0.0142 & 0.0133 & 0.0106 & 0.0141 \\
    & User mean F1 & \textbf{0.101} & 0.186 & 0.275 & 0.377 & 0.552 & 0.596 & 0.649 & 0.683 & 0.917 \\
    &SD& 0.00886 & 0.00745 & 0.0104 & 0.0251 & 0.00357 & 0.00650 & 0.0150 & 0.0274 & 0.0107 \\
  \hline
  Hand & Activity mean F1 & 0.246 & 0.551 & 0.636 & 0.718 & 0.725 & 0.744 & 0.749 & 0.760 & \textbf{0.762} \\
  Chest       &SD& 0.00902 & 0.00636 & 0.00953 & 0.00896 & 0.0249 & 0.0108 & 0.0120 & 0.0276 & 0.0184 \\
         & User mean F1 & \textbf{0.0786} & 0.239 & 0.455 & 0.714 & 0.827 & 0.896 & 0.911 & 0.942 & 0.986 \\
         &SD& 0.0346 & 0.0493 & 0.00441 & 0.00876 & 0.00737 & 0.00370 & 0.00603 & 0.00490 & 0.00377 \\
  \hline
  Chest & Activity mean F1 & 0.253 & 0.530 & 0.605 & 0.653 & 0.667 & 0.671 & 0.685 & 0.685 & \textbf{0.687} \\
  Ankle     &SD& 0.0132 & 0.0106 & 0.0127 & 0.0167 & 0.0191 & 0.0179 & 0.0120 & 0.0116 & 0.0186 \\
          & User mean F1 & \textbf{0.0755} & 0.214 & 0.320 & 0.556 & 0.680 & 0.755 & 0.791 & 0.832 & 0.967 \\
          &SD& 0.0235 & 0.0286 & 0.00836 & 0.00938 & 0.00745 & 0.00142 & 0.0122 & 0.0191 & 0.00460 \\
  \hline
  Ankle & Activity mean F1 & 0.259 & 0.565 & 0.663 & 0.719 & 0.730 & 0.724 & 0.734 & 0.737 & \textbf{0.803} \\
  Hand       &SD& 0.0228 & 0.00473 & 0.00485 & 0.00706 & 0.00540 & 0.00903 & 0.00454 & 0.0131 & 0.0126 \\
         & User mean F1 & \textbf{0.0748} & 0.278 & 0.431 & 0.693 & 0.813 & 0.868 & 0.898 & 0.912 & 0.987 \\
         &SD& 0.0259 & 0.0151 & 0.0133 & 0.00509 & 0.00245 & 0.00305 & 0.00713 & 0.00331 & 0.00250 \\
  \hline
  All Sensors & Activity mean F1 & 0.329 & 0.641 & 0.705 & 0.767 & 0.750 & 0.765 & 0.772 & 0.773 & \textbf{0.799} \\
        &SD& 0.0326 & 0.00622 & 0.00498 & 0.00736 & 0.0146 & 0.0106 & 0.0156 & 0.0132 & 0.0183 \\
        & User mean F1 & \textbf{0.0882} & 0.282 & 0.467 & 0.742 & 0.852 & 0.916 & 0.935 & 0.943 & 0.990 \\
        &SD& 0.0144 & 0.0392 & 0.0240 & 0.0181 & 0.00623 & 0.00377 & 0.00291 & 0.00236 & 0.00188 \\
  \hline
 \end{tabular}
 \end{adjustbox}
\end{table}

% Table 7
\begin{table*}[t]
\renewcommand{\arraystretch}{1.3}
\caption{Experimental Results for MotionSense (Estimator: CNN). SD = Standard Deviation.}
 \label{table:motionsense}
 \centering
 \scalebox{0.9}{
 \begin{tabular}{ccccc}
  \hline
  Target & Metric & Baseline & AAE~\cite{bib:AAE} & C-AAE (Proposed) \\
  \hline \hline
  Activity & Mean F1 & 0.903 & 0.883 & \textbf{0.874} \\
       &SD & 0.0180 & 0.0119 & 0.0134 \\
  \hline
  User & Mean F1 & 0.532 & 0.259 & \textbf{0.0732} \\
    &SD & 0.0188 & 0.0171 & 0.0104 \\
  \hline
 \end{tabular}
 }
\end{table*}

\subsection{Evaluation Results}

Building on the findings from the preliminary evaluation, we now present the results of the comparative experiments involving the proposed method, C-AAE, and the prior approach, AAE~\cite{bib:AAE}. These experiments were conducted based on the Application Criteria described earlier, focusing on both activity recognition performance (Requirement 1) and user anonymity (Requirement 2).

\subsubsection{Results for MotionSense}
\label{subsubsec:motionsense}

The activity and user recognition accuracies of AAE and C-AAE using MotionSense as input data for CNN are shown in Table~\ref{table:motionsense}. Here, the baseline values represent the results when the input data are directly fed into the estimator without any anonymization. As a result, both AAE and C-AAE maintained accuracy for activity recognition, satisfying Application Criterion 1. Next, for user recognition, Application Criterion 2 for MotionSense requires the user recognition accuracy to be below approximately 4.17\%. The results for the prior method, AAE, show an accuracy of approximately 25.9\%, indicating that introducing static activity labels (std, sit) into the experimental dataset prevented Criterion 2 from being achieved.
On the other hand, the proposed method, C-AAE, achieved a user recognition accuracy of approximately 7.32\%, reducing the accuracy by about 18\% compared to AAE. However, it still did not meet Criterion 2. In summary, when using MotionSense as the input data, C-AAE outperformed AAE in reducing privacy risk while satisfying Application Criterion 1. However, it did not satisfy Criterion 2. Nevertheless, compared to the trade-offs between application accuracy and privacy risk observed in the differential privacy methods discussed in Section~\ref{subsec:dp}, C-AAE significantly mitigated these trade-offs.

\subsubsection{Results for PAMAP2}
\label{subsubsec:pamap2}

For PAMAP2, the results were evaluated in two scenarios: using data from a single IMU (hand, chest, or ankle) and using a combination of multiple IMUs. First, Table~\ref{table:pamap2-result} shows the results when CNN was used as the estimator with data from a single IMU. The results indicate that, in all cases, the user recognition accuracy was reduced by approximately 7\% to 10\% compared to the baseline and AAE. For activity recognition, the accuracy improved in the chest and ankle IMUs cases compared to AAE. In the case of the hand IMU, considering the standard deviation of AAE, the difference is negligible. Overall, the proposed method demonstrates that activity recognition can be performed with reduced privacy risk compared to the conventional method. In other words, compared to the differential privacy methods in Section~\ref{subsec:dp}, the proposed method resolves the trade-offs between application accuracy and privacy risk.

Furthermore, to satisfy Application Criterion 2 for PAMAP2, the user recognition accuracy must be below approximately 11.1\%. This criterion was achieved in the cases of the hand and ankle IMUs. Next, regarding Application Criterion 1, it was satisfied only in the case of the ankle IMU. Thus, the results indicate that both criteria were met in this scenario.

% Table 8
\begin{table*}[t]
\renewcommand{\arraystretch}{1.3}
 \caption{Experimental Results Using Single IMU Data from PAMAP2 (Estimator: CNN)}
 \label{table:pamap2-result}
 \centering
  \scalebox{0.9}{
 \begin{tabular}{cccccc}
  \hline
  IMU Location & Target & Metric & Baseline & AAE~\cite{bib:AAE} & C-AAE (Proposed) \\
  \hline \hline
  Hand & Activity & Mean F1 & \textbf{0.712} & 0.686 & 0.656 \\
    & &Standard Deviation & 0.00765 & 0.0165 & 0.00645 \\
    & User & Mean F1 & 0.914 & 0.203 & \textbf{0.102} \\
    &    &Standard Deviation & 0.0106 & 0.0179 & 0.0145 \\
  \hline
  Chest & Activity & Mean F1 & \textbf{0.662} & 0.564 & 0.604 \\
    & &Standard Deviation & 0.0257 & 0.0148 & 0.0159 \\
    & User & Mean F1 & 0.918 & 0.206 & \textbf{0.139} \\
    &    &Standard Deviation & 0.00936 & 0.0182 & 0.0207 \\
  \hline
  Ankle & Activity & Mean F1 & \textbf{0.676} & 0.636 & 0.643 \\
    & &Standard Deviation & 0.0141 & 0.0244 & 0.00738 \\
    & User & Mean F1 & 0.917 & 0.139 & \textbf{0.0504} \\
    &    &Standard Deviation & 0.0107 & 0.0227 & 0.0155 \\
  \hline
 \end{tabular}
 }
\end{table*}

% Table 9
\begin{table*}[t]
\renewcommand{\arraystretch}{1.3}
 \caption{Experimental Results Using Multiple IMU Data from PAMAP2 (Estimator: CNN)}
 \label{table:pamap2-result1}
 \centering
  \scalebox{0.9}{
 \begin{tabular}{cccccc}
  \hline
  Input Pattern & Target & Metric & Baseline & AAE~\cite{bib:AAE} & C-AAE (Proposed) \\
  \hline \hline
  Input 1 & Activity & Mean F1 & \textbf{0.762} & 0.715 & 0.601 \\
  (Hand and Chest IMUs) & &Standard Deviation & 0.0184 & 0.00877 & 0.00756 \\
    & User & Mean F1 & 0.986 & 0.184 & \textbf{0.115} \\
    &    &Standard Deviation & 0.00377 & 0.0260 & 0.0119 \\
  \hline
  Input 2 & Activity & Mean F1 & \textbf{0.687} & 0.664 & 0.643 \\
  (Chest and Ankle IMUs) & &Standard Deviation & 0.0186 & 0.0148 & 0.0159 \\
    & User & Mean F1 & 0.967 & 0.132 & \textbf{0.0552} \\
    &    &Standard Deviation & 0.00460 & 0.0182 & 0.0207 \\
  \hline
  Input 3 & Activity & Mean F1 & \textbf{0.803} & 0.767 & 0.780 \\
  (Hand and Ankle IMUs) & &Standard Deviation & 0.0126 & 0.0128 & 0.00718 \\
    & User & Mean F1 & 0.987 & 0.237 & \textbf{0.139} \\
    &    &Standard Deviation & 0.00250 & 0.0181 & 0.0101 \\
  \hline
  Input 4 & Activity & Mean F1 & \textbf{0.799} & 0.692 & 0.700 \\
  (All Three IMUs) & &Standard Deviation & 0.0183 & 0.0115 & 0.0111 \\
    & User & Mean F1 & 0.990 & 0.250 & \textbf{0.157} \\
    &    &Standard Deviation & 0.00188 & 0.0237 & 0.0137 \\
  \hline
 \end{tabular}
 }
\end{table*}

% Table 10
\begin{table}[t]
\renewcommand{\arraystretch}{1.3}
 \caption{Experimental Results Using Multiple IMU Data from PAMAP2 \\ (Estimator: LGBM, Metric: Mean F1)}
 \label{table:pamap2-result2-lgbm}
 \centering
\scalebox{0.9}{
 \begin{tabular}{cccccc}
  \hline
  Input Pattern & Target & Baseline & AAE~\cite{bib:AAE} & C-AAE (Proposed) \\
  \hline \hline
  Input 1 & Activity & \textbf{0.905} & 0.916 & 0.847 \\
      & User   & 0.613 & 0.460 & \textbf{0.281} \\
  \hline
  Input 2 & Activity & 0.876 & \textbf{0.878} & 0.806 \\
      & User   & 0.653 & 0.390 & \textbf{0.249} \\
  \hline
  Input 3 & Activity & \textbf{0.895} & 0.888 & 0.797 \\
      & User   & 0.670 & 0.369 & \textbf{0.223} \\
  \hline
  Input 4 & Activity & 0.914 & \textbf{0.921} & 0.894 \\
      & User   & 0.717 & 0.460 & \textbf{0.323} \\
  \hline
 \end{tabular}
 }
\end{table}

% Table 11
\begin{table}[t]
\renewcommand{\arraystretch}{1.3}
 \caption{Experimental Results Using Multiple IMU Data from PAMAP2 \\ (Estimator: RF, Metric: Mean F1)}
 \label{table:pamap2-result2-rf}
 \centering
\scalebox{0.9}{
 \begin{tabular}{cccccc}
  \hline
  Input Pattern & Target & Baseline & AAE~\cite{bib:AAE} & C-AAE (Proposed) \\
  \hline \hline
  Input 1 & Activity & 0.903 & \textbf{0.915} & 0.839 \\
      & User   & 0.621 & 0.482 & \textbf{0.296} \\
  \hline
  Input 2 & Activity & \textbf{0.883} & 0.881 & 0.791 \\
      & User   & 0.637 & 0.402 & \textbf{0.262} \\
  \hline
  Input 3 & Activity & 0.897 & \textbf{0.899} & 0.779 \\
      & User   & 0.653 & 0.381 & \textbf{0.236} \\
  \hline
  Input 4 & Activity & 0.916 & \textbf{0.921} & 0.884 \\
      & User   & 0.689 & 0.454 & \textbf{0.321} \\
  \hline
 \end{tabular}
 }
\end{table}

% Table 12
\begin{table}[t]
\renewcommand{\arraystretch}{1.3}
 \caption{Experimental Results Using Multiple IMU Data from PAMAP2 \\ (Estimator: LR, Metric: Mean F1)}
 \label{table:pamap2-result2-lr}
 \centering
\scalebox{0.9}{
 \begin{tabular}{cccccc}
  \hline
  Input Pattern & Target & Baseline & AAE~\cite{bib:AAE} & C-AAE (Proposed) \\
  \hline \hline
  Input 1 & Activity & 0.881 & \textbf{0.898} & 0.811 \\
      & User   & 0.449 & 0.226 & \textbf{0.153} \\
  \hline
  Input 2 & Activity & 0.832 & \textbf{0.855} & 0.790 \\
      & User   & 0.457 & \textbf{0.152} & 0.168 \\
  \hline
  Input 3 & Activity & 0.863 & \textbf{0.880} & 0.779 \\
      & User   & 0.502 & \textbf{0.138} & 0.142 \\
  \hline
  Input 4 & Activity & 0.889 & \textbf{0.906} & 0.872 \\
      & User   & 0.558 & 0.240 & \textbf{0.193} \\
  \hline
 \end{tabular}
 }
\end{table}

% Figure 3 ― two confusion-matrix images side by side (ACM single-column review format)
\begin{figure}[t]
 \centering
 %
 %---- left sub-figure --------------------------------------------------
 \begin{subfigure}{0.45\columnwidth}  % 0.48 to leave a small gutter
  \centering
  \includegraphics[width=\linewidth]{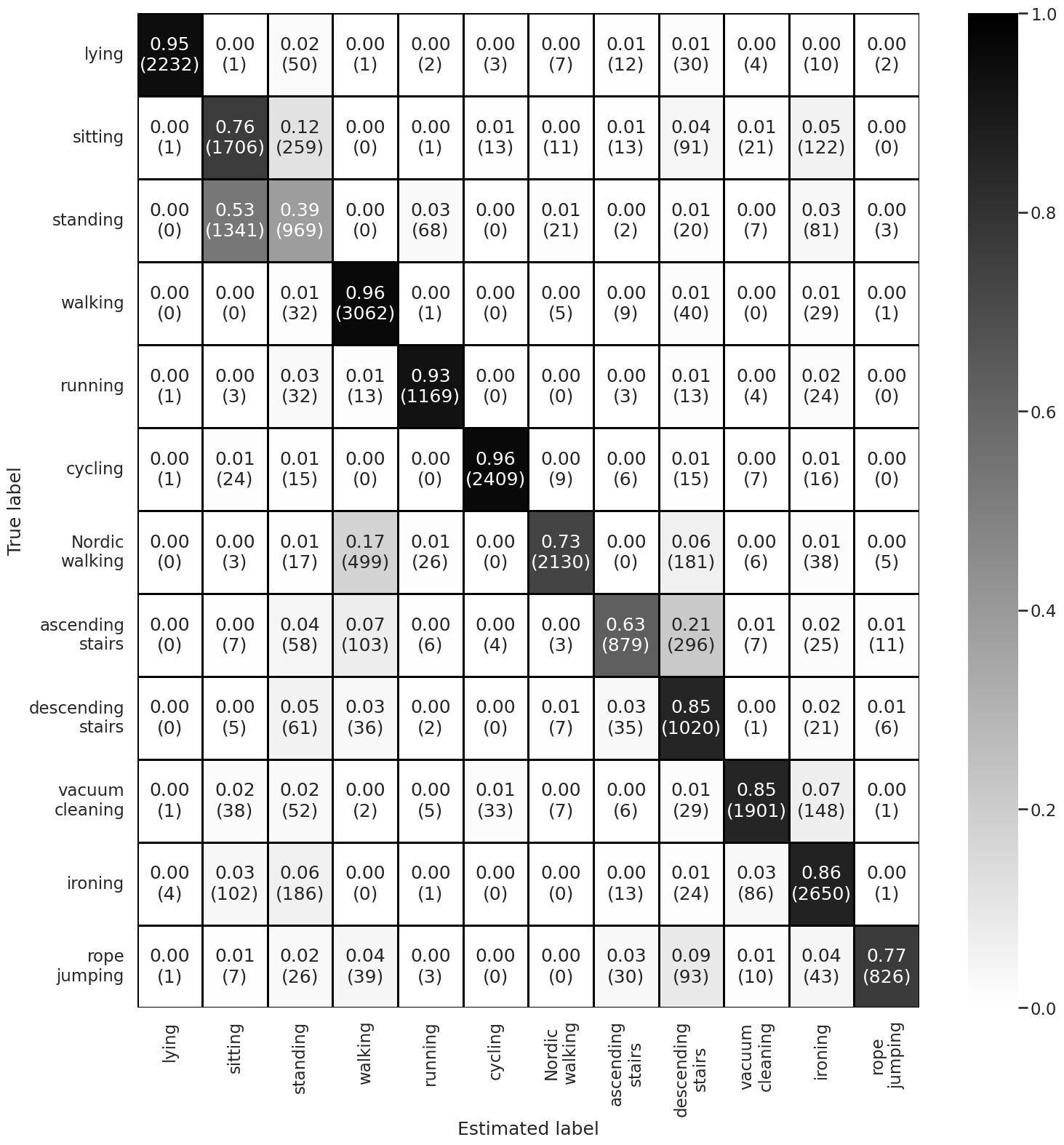}
  \caption{Baseline (Input 3)}
  \label{fig:act-result3-a}
 \end{subfigure}
 \hfill                  % horizontal space between the two figures
 %
 %---- right sub-figure -------------------------------------------------
 \begin{subfigure}{0.45\columnwidth}
  \centering
  \includegraphics[width=\linewidth]{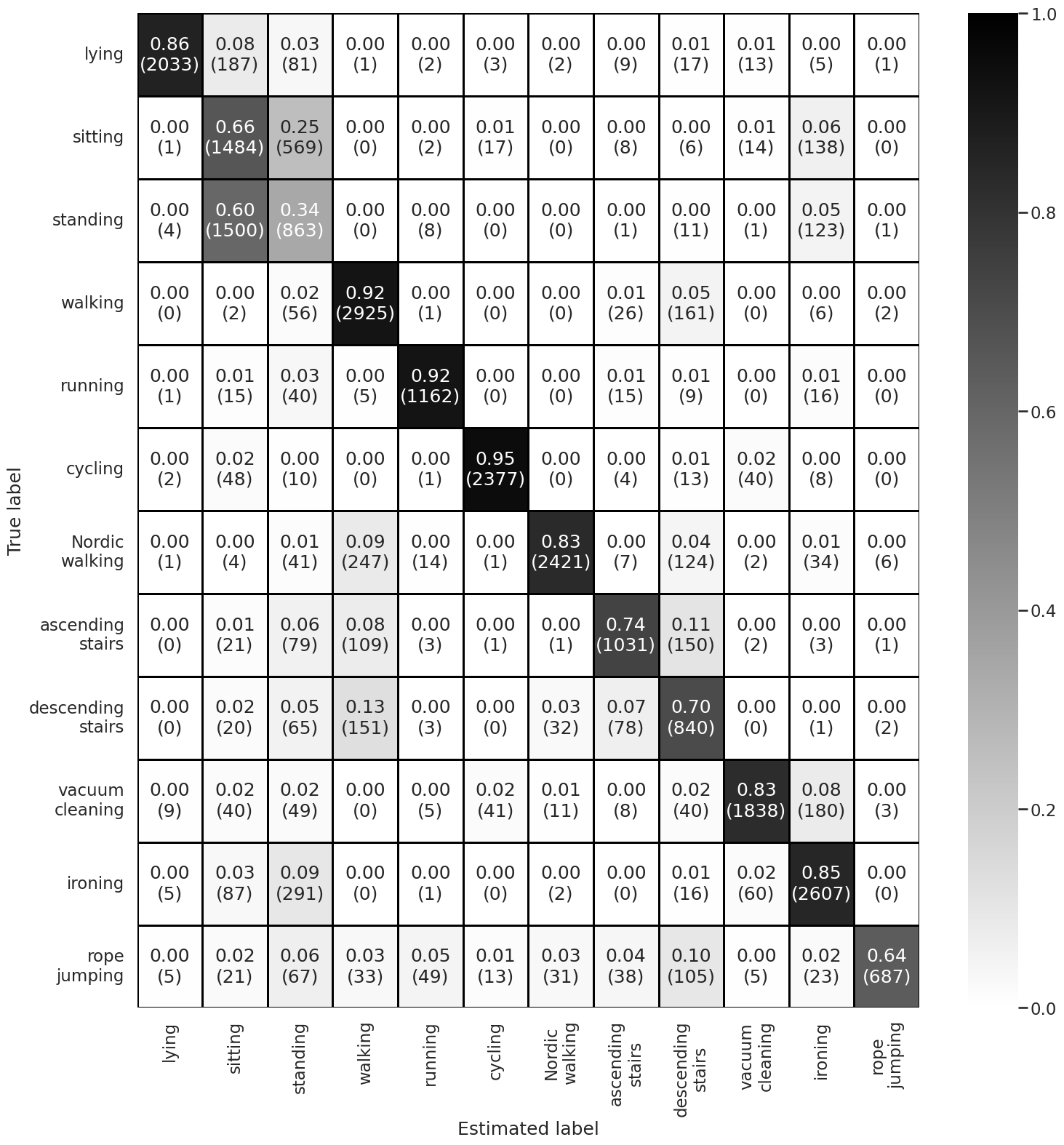}
  \caption{C-AAE (Input 3)}
  \label{fig:act-result3-b}
 \end{subfigure}
 \caption{Confusion matrices for activity recognition on PAMAP2 using multiple-IMU data (estimator: CNN).}
 \label{fig:act-result-pamap2}
\end{figure}

% Figure 4 — two confusion-matrix plots side by side (ACM single-column)
\begin{figure}[t]
 \centering
 \begin{subfigure}{0.45\columnwidth}
  \centering
  \includegraphics[width=\linewidth]{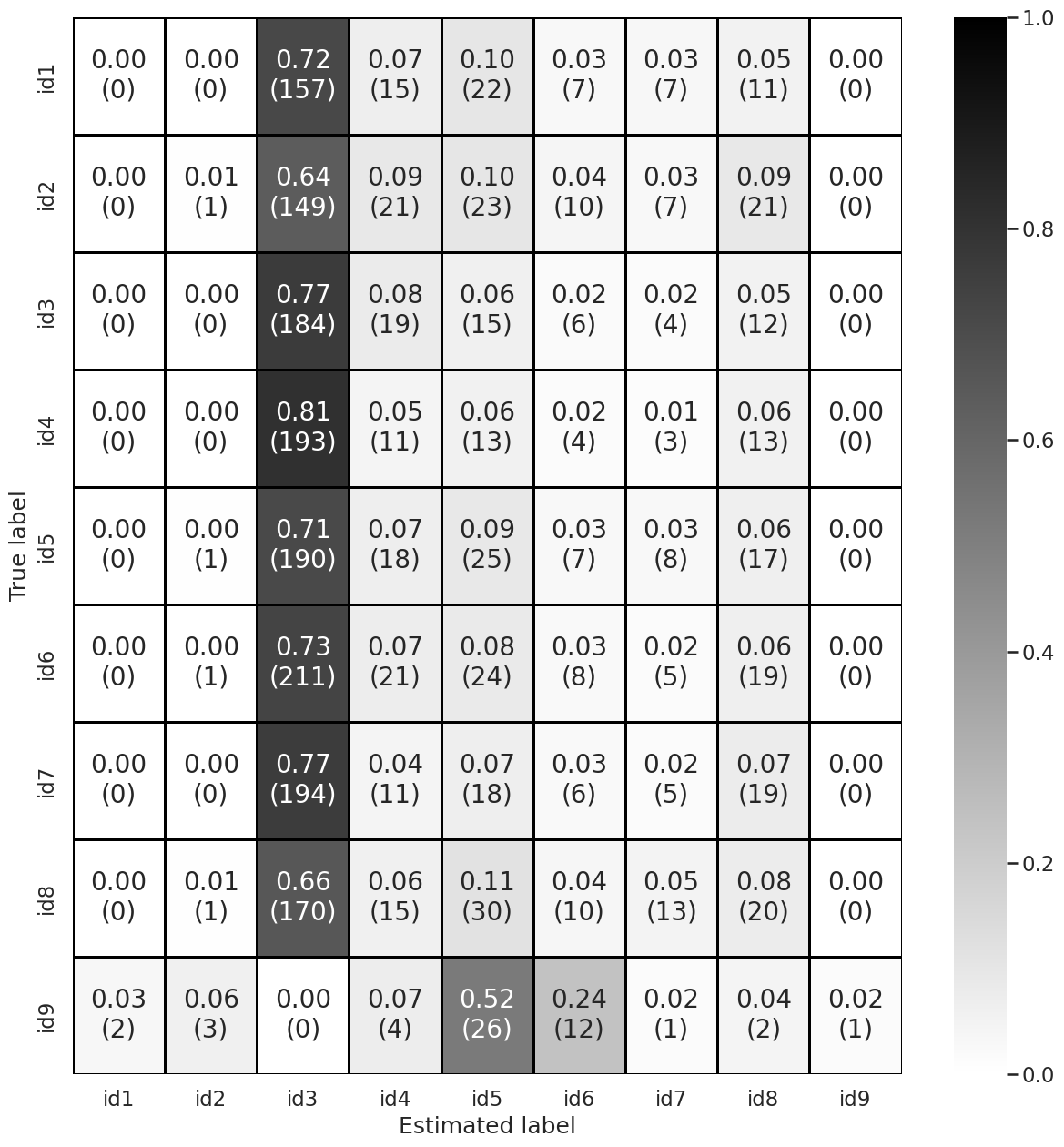}
  \caption{Input 2}
  \label{fig:result3-a}
 \end{subfigure}
 \hfill
 \begin{subfigure}{0.45\columnwidth}
  \centering
  \includegraphics[width=\linewidth]{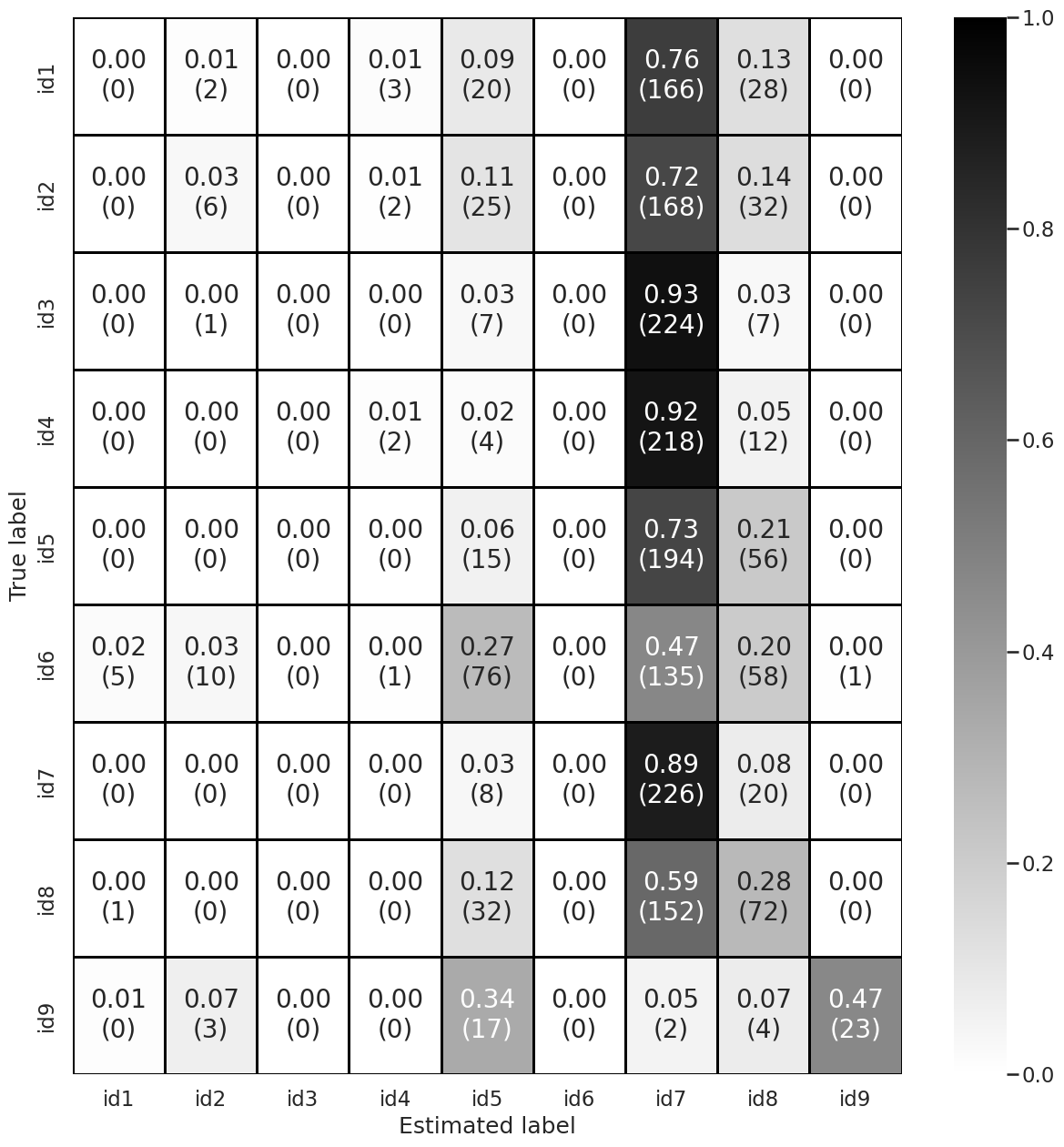}
  \caption{Input 3}
  \label{fig:result3-b}
 \end{subfigure}
 \caption{Confusion matrices for user-recognition accuracy on PAMAP2 using multiple-IMU data (estimator: CNN).}
 \label{fig:user-result-pamap2}
\end{figure}

Next, we evaluate the case where combinations of multiple IMU data from PAMAP2 are used as input to the estimator. The input data patterns are categorized into the following four types:\\\\
(Input 1) Using IMUs from the hand and chest\\
(Input 2) Using IMUs from the chest and ankle\\
(Input 3) Using IMUs from the hand and ankle\\
(Input 4) Using IMUs from all three locations\\

The results for each input data pattern are summarized for both CNN and machine learning algorithms as estimators. First, the results for CNN are shown in Table~\ref{table:pamap2-result1}. As a result, for Inputs 2 and 3, C-AAE successfully reduced user recognition accuracy significantly while maintaining activity recognition accuracy at the baseline level. For Input 4, while there was no substantial difference in activity recognition accuracy between C-AAE and AAE, the accuracy decreased by approximately 10\% compared to the baseline. However, C-AAE proved to be more effective than AAE in reducing user recognition accuracy. Finally, for Input 1, although user recognition accuracy was reduced compared to AAE, activity recognition accuracy fell significantly below both the baseline and AAE. In conclusion, for input patterns other than Input 1, the trade-offs observed in the differential privacy methods discussed in Section~\ref{subsec:dp} were resolved.

In PAMAP2, to satisfy Application Criterion 2, user recognition accuracy must be below approximately 11.1\%. Based on the evaluation following the application criteria, Input 2 and Input 3 meet Criterion 1, while only Input 2 meets Criterion 2. However, for Inputs 1, 3, and 4, while Criterion 2 was not satisfied, the results were close to the threshold. Figure~\ref{fig:act-result-pamap2} shows a comparison of confusion matrices for activity recognition accuracy between the baseline and the proposed method for Input 3. Figure~\ref{fig:user-result-pamap2} presents the confusion matrices for user recognition accuracy using C-AAE for Inputs 2 and 3. In Figure~\ref{fig:act-result-pamap2}, there is almost no difference between the baseline and the proposed method, with some activity types showing improved recognition accuracy using the proposed method. This demonstrates that Criterion 1 is satisfied based on the confusion matrices. Next, in Figure~\ref{fig:user-result-pamap2}, the results for Input 2 (Figure~\ref{fig:result3-a}) show that predictions are concentrated on user id3, with no other users being identified. Thus, the criterion of users being identified only at a random probability is satisfied. However, for Input 3 (Figure~\ref{fig:result3-b}), which does not satisfy Criterion 2, the results are similar to those of Input 2, with predictions concentrated on user id7. Additionally, a few instances of user id7 and id9 being identified indicate that Criterion 2 is not fully satisfied.

Next, the results using machine learning algorithms as estimators are shown in Tables~\ref{table:pamap2-result2-lgbm}, \ref{table:pamap2-result2-rf}, and \ref{table:pamap2-result2-lr}. For LGBM and RF as estimators, the results show almost no differences between the two. Compared to the baseline, Input 4 maintained activity recognition accuracy, while the other inputs showed a decrease of approximately 6–10\%. For user recognition accuracy, C-AAE successfully reduced accuracy by approximately 13–19\% for all inputs compared to AAE. For LR as an estimator, similar to the other estimators, Input 4 maintained activity recognition accuracy compared to the baseline, while the other inputs showed a decrease of approximately 4–9\%. Regarding user recognition accuracy, Inputs 1 and 4 achieved a reduction of approximately 5–7\% compared to AAE, but Inputs 2 and 3 showed increased recognition accuracy. The reduction in user recognition accuracy for Inputs 1 and 4 was almost proportional to the decrease in activity recognition accuracy, indicating that the effect of C-AAE was insignificant. Based on the application criteria evaluation, Input 4 satisfied Criterion 1 for all machine learning algorithms, but none of the inputs satisfied Criterion 2.

Overall, when using PAMAP2 as input data, the effectiveness of C-AAE was not consistently observed across all input data patterns and estimators. Therefore, while the application criteria were not fully satisfied in most cases, it can be concluded that C-AAE successfully reduced user recognition accuracy to some extent while maintaining activity recognition accuracy.

\section{Discussion and Future Work}
\label{sec:consideration}

\begin{figure}[t]
 \centering
 % ---------- Row 1 : Activity ----------------------------------------
 \begin{subfigure}{0.32\columnwidth}
  \centering
  \includegraphics[width=\linewidth]{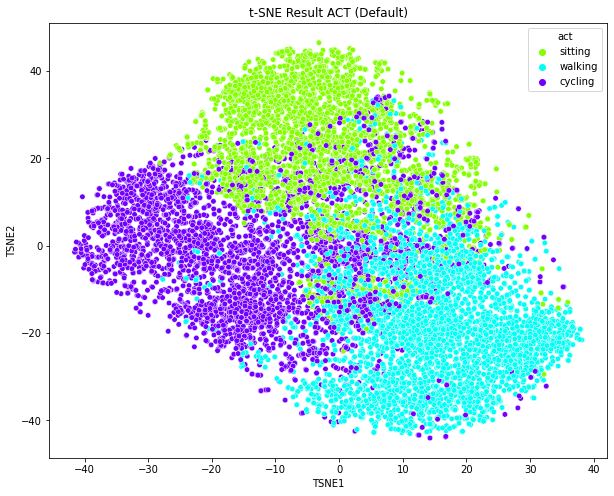}
  \caption{Activity (raw)}
  \label{fig:result4-a}
 \end{subfigure}\hfill
 \begin{subfigure}{0.32\columnwidth}
  \centering
  \includegraphics[width=\linewidth]{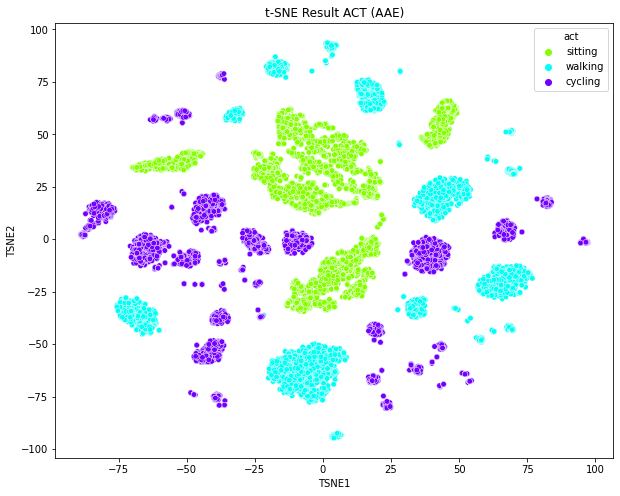}
  \caption{Activity (AAE)}
   \label{fig:result4-c}
 \end{subfigure}\hfill
 \begin{subfigure}{0.32\columnwidth}
  \centering
  \includegraphics[width=\linewidth]{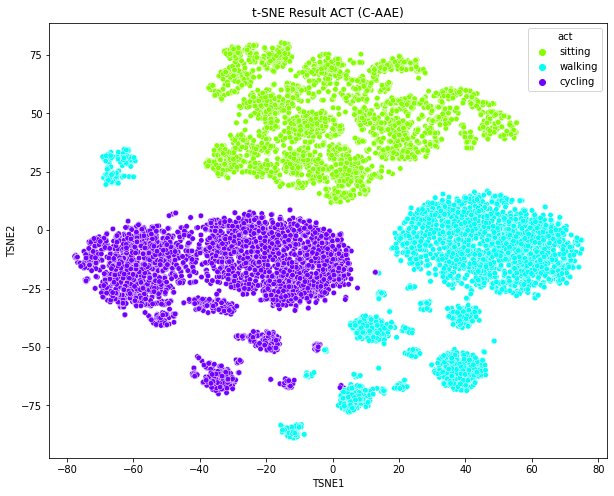}
  \caption{Activity (C-AAE)}
  \label{fig:result4-e}
 \end{subfigure}

 \vspace{0.8em} % vertical gap between rows

 % ---------- Row 2 : User ID -----------------------------------------
 \begin{subfigure}{0.32\columnwidth}
  \centering
  \includegraphics[width=\linewidth]{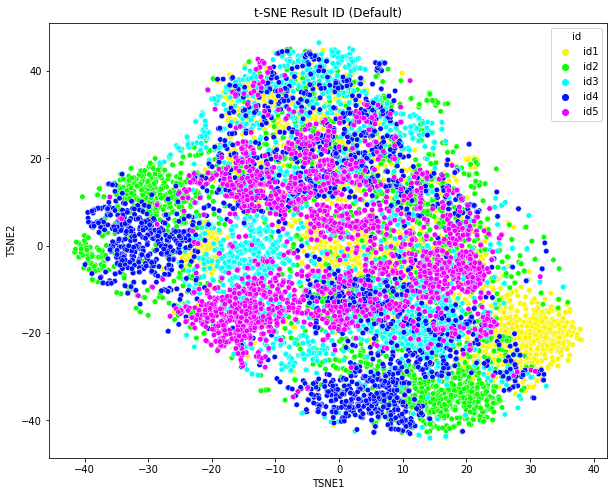}
  \caption{User ID (raw)}
  \label{fig:result4-b}
 \end{subfigure}\hfill
 \begin{subfigure}{0.32\columnwidth}
  \centering
  \includegraphics[width=\linewidth]{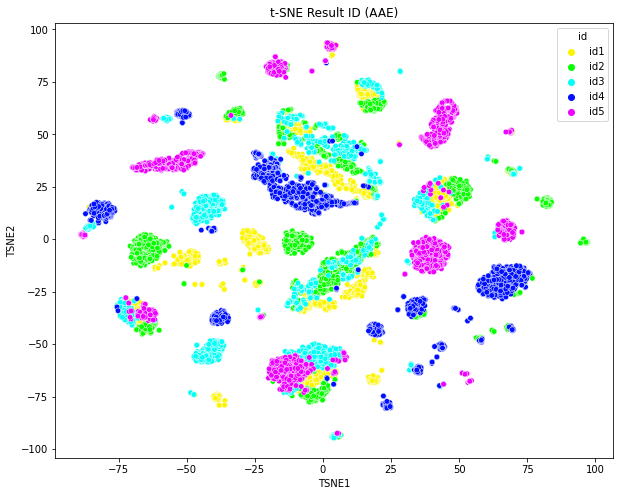}
  \caption{User ID (AAE)}
   \label{fig:result4-d}
 \end{subfigure}\hfill
 \begin{subfigure}{0.32\columnwidth}
  \centering
  \includegraphics[width=\linewidth]{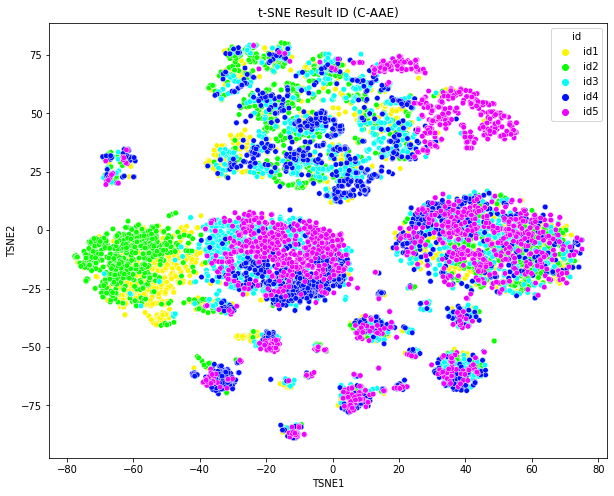}
  \caption{User ID (C-AAE)}
  \label{fig:result4-f}
 \end{subfigure}

 \caption{t-SNE visualisations of latent representations. 
 Top row: activity-label views for Raw, AAE, and proposed C-AAE. 
 Bottom row: corresponding user-ID views.}
 \label{fig:t-sne}
\end{figure}

Based on the evaluation results in Section~\ref{sec:evaluation}, it was found that C-AAE is a data transformation method capable of reducing privacy risks even when the number of target activities is increased or when different estimators are used, compared to the AAE method in previous research~\cite{bib:AAE}. Figure~\ref{fig:t-sne} shows the results of dimensionality reduction using t-SNE for a subset of the input data (PAMAP2 with three activity labels and five user ID labels), visualized by activity and user ID labels. Figures~\ref{fig:result4-a} and \ref{fig:result4-b} represent the results without any anonymization, Figures~\ref{fig:result4-c} and \ref{fig:result4-d} show the results when AAE was applied to the input data, and Figures~\ref{fig:result4-e} and \ref{fig:result4-f} present the results when the proposed C-AAE was applied. From these results, when visualized by activity labels (Figures~\ref{fig:result4-a}, \ref{fig:result4-c}, and \ref{fig:result4-e}), clusters are approximately formed for each activity label. This clustering tendency is particularly strong in the cases without anonymization and with the proposed method, suggesting that activity recognition maintained high data utility without significant degradation in accuracy even after anonymization. When visualized by user ID labels (Figures~\ref{fig:result4-b}, \ref{fig:result4-d}, and \ref{fig:result4-f}), in the case without anonymization (Figure~\ref{fig:result4-b}), while there is overlap of multiple ID labels in the central region, several clusters specific to individual labels are also observed. Next, with AAE-based anonymization (Figure~\ref{fig:result4-d}), clusters are scattered across multiple regions, and the data for each label appears slightly dispersed. Compared to Figure~\ref{fig:result4-b}, there is a stronger tendency for clusters to overlap multiple ID labels. Finally, with the proposed C-AAE anonymization (Figure~\ref{fig:result4-f}), the data for each label is more scattered compared to Figures~\ref{fig:result4-b} and \ref{fig:result4-d}, making it difficult for clusters to form around single ID labels. Additionally, compared to the results in Figure~\ref{fig:result4-d}, the overall distribution suggests a further reduction in recognition accuracy. In particular, the data for static activity labels shows a strong tendency to overlap across multiple ID labels, indicating that C-AAE is more effective at anonymizing static activities. This explains why the user recognition accuracy for each input pattern in the proposed method's evaluation results decreased compared to AAE.

Next, we discuss the evaluation results when using machine learning algorithms. In the present evaluation, the application criterion of requirement 2 was not satisfied when machine learning algorithms were used. This is likely due to the presence or absence of feature extraction. The data input to CNN consisted of the magnitude of the 3-axis acceleration and gyroscope data, to which AAE and then ADPCM were applied. On the other hand, when inputting to machine learning algorithms, AAE and ADPCM were applied to the raw 3-axis acceleration and gyroscope data before feature extraction was performed. As a result, the amount of information available for user identification increased due to feature extraction, making it easier for machine learning algorithms to recognize users.
Additionally, when ADPCM compression was performed after feature extraction, there was little difference in the user recognition accuracy reduction rate. To fully protect the privacy of sensor data, it is necessary to suppress user identification risks regardless of the learning algorithm or data format (conversely, activity recognition should achieve sufficient accuracy with anonymized data regardless of the learning algorithm used). However, as far as the authors are aware, prior research on privacy protection technologies has not investigated whether privacy can still be guaranteed when the input data format is changed. Therefore, future work will investigate the range of adaptable data formats for anonymization.

Furthermore, additional improvements are needed to implement C-AAE into an application to achieve requirement 2. In this regard, the TEMPDIFF framework~\cite{bib:har-2}, which uses the differential privacy method Private Aggregation of Teacher Ensembles (PATE)~\cite{bib:pate}, is considered a promising reference. TEMPDIFF is a learning method that leverages the temporal characteristics of wearable device data, and prior research has demonstrated its success in reducing privacy risks while maintaining activity recognition accuracy. By combining such methods with C-AAE, it is believed that privacy risks can be further mitigated.

Based on the evaluation results, AAE was found to reduce privacy risks for certain activity patterns effectively but failed to achieve complete anonymization when the number of target activity patterns increased. The utility of anonymization varies depending on the application specifications, such as which activities are targeted and the required activity recognition accuracy. However, the proposed C-AAE successfully addressed some of these challenges, making it a more versatile anonymization method. Additionally, a secondary benefit of C-AAE is its ability to reduce the data volume processed on the server and the communication data volume between edge devices and the server by approximately 75\% through performing encoding directly on the user device. In the future, we plan to implement C-AAE in applications that utilize activity recognition and analyze the user experience it provides.

\section{Conclusion}
\label{sec:conclusion}

In this study, we focused on the challenges of AAE~\cite{bib:AAE}, a method developed in prior research, which struggles to achieve complete anonymization when the variety of target activities increases or when different estimators are used. To address these challenges, we defined the criteria that privacy-aware activity recognition applications should meet, and proposed and evaluated C-AAE as a data anonymization method to fulfill these criteria. The evaluation results showed that, in experiments using both the MotionSense and PAMAP2 datasets, C-AAE successfully maintained the activity recognition accuracy criteria (requirement 1) in most cases while significantly reducing privacy risks (user identification risks) that could arise with AAE. However, there were few cases where the user recognition accuracy criteria (requirement 2) were fully satisfied. Nevertheless, the results indicate that in many cases, the performance was close to meeting requirement 2. 
Through this study, we demonstrated that C-AAE has the potential to enable applications with minimal privacy risks while reducing the data usage, which is not achievable with AAE. Moving forward, we aim to improve C-AAE to make it a more practical privacy-preserving technology. We also plan to integrate C-AAE into sensor-based applications, analyze user experiences, and extend the framework to support a wider variety of datasets. 
While our evaluation focused on MotionSense and PAMAP2, which are commonly used in activity recognition research, future work should explore the applicability of C-AAE to other publicly available datasets such as MobiAct~\cite{vavoulas2016mobiact}, KU-HAR~\cite{sikder2021ku}, and Opportunity++~\cite{ciliberto2021opportunity++}. These datasets offer larger participant pools, a broader range of activity types, and more complex sensing environments, including multimodal inputs. Evaluating C-AAE under such conditions will be essential to confirm its robustness and generalizability in more realistic and varied usage scenarios.

% \vspace{-4mm}
% \section*{Acknowledgment}
% This work was supported by JST, PRESTO Grant Number JPMJPR21P7, Japan.
% \vspace{-3mm}

\bibliographystyle{ACM-Reference-Format}
\bibliography{ref}


\begin{thebibliography}{34}

%%% the Web address.

\ifx \showCODEN    \undefined \def \showCODEN     #1{\unskip}     \fi
\ifx \showDOI      \undefined \def \showDOI       #1{#1}\fi
\ifx \showISBNx    \undefined \def \showISBNx     #1{\unskip}     \fi
\ifx \showISBNxiii \undefined \def \showISBNxiii  #1{\unskip}     \fi
\ifx \showISSN     \undefined \def \showISSN      #1{\unskip}     \fi
\ifx \showLCCN     \undefined \def \showLCCN      #1{\unskip}     \fi
\ifx \shownote     \undefined \def \shownote      #1{#1}          \fi
\ifx \showarticletitle \undefined \def \showarticletitle #1{#1}   \fi
\ifx \showURL      \undefined \def \showURL       {\relax}        \fi
% invisible to TeX
\providecommand\bibfield[2]{#2}
\providecommand\bibinfo[2]{#2}
\providecommand\natexlab[1]{#1}
\providecommand\showeprint[2][]{arXiv:#2}

\bibitem[Allgaier and Pryss(2024)]%
        {bib:skcv}
\bibfield{author}{\bibinfo{person}{Johannes Allgaier} {and} \bibinfo{person}{Rüdiger Pryss}.} \bibinfo{year}{2024}\natexlab{}.
\newblock \showarticletitle{Cross-Validation Visualized: A Narrative Guide to Advanced Methods}.
\newblock \bibinfo{journal}{\emph{Machine Learning and Knowledge Extraction}} \bibinfo{volume}{6}, \bibinfo{number}{2} (\bibinfo{year}{2024}), \bibinfo{pages}{1378--1388}.
\newblock
\showISSN{2504-4990}
\urldef\tempurl%
\url{https://doi.org/10.3390/make6020065}
\showDOI{\tempurl}


\bibitem[Bevilacqua et~al\mbox{.}(2019)]%
        {bib:cnn-2}
\bibfield{author}{\bibinfo{person}{Antonio Bevilacqua}, \bibinfo{person}{Kyle MacDonald}, \bibinfo{person}{Aamina Rangarej}, \bibinfo{person}{Venessa Widjaya}, \bibinfo{person}{Brian Caulfield}, {and} \bibinfo{person}{Tahar Kechadi}.} \bibinfo{year}{2019}\natexlab{}.
\newblock \showarticletitle{Human Activity Recognition with Convolutional Neural Networks}. In \bibinfo{booktitle}{\emph{Machine Learning and Knowledge Discovery in Databases}}, \bibfield{editor}{\bibinfo{person}{Ulf Brefeld}, \bibinfo{person}{Edward Curry}, \bibinfo{person}{Elizabeth Daly}, \bibinfo{person}{Brian MacNamee}, \bibinfo{person}{Alice Marascu}, \bibinfo{person}{Fabio Pinelli}, \bibinfo{person}{Michele Berlingerio}, {and} \bibinfo{person}{Neil Hurley}} (Eds.). \bibinfo{publisher}{Springer International Publishing}, \bibinfo{address}{Cham}, \bibinfo{pages}{541--552}.
\newblock
\showISBNx{978-3-030-10997-4}


\bibitem[Bigelli et~al\mbox{.}(2024)]%
        {bib:MLP-AAE}
\bibfield{author}{\bibinfo{person}{L. Bigelli}, \bibinfo{person}{C. Contoli}, \bibinfo{person}{V. Freschi}, {and} \bibinfo{person}{E. Lattanzi}.} \bibinfo{year}{2024}\natexlab{}.
\newblock \showarticletitle{Privacy preservation in sensor-based Human Activity Recognition through autoencoders for low-power IoT devices}.
\newblock \bibinfo{journal}{\emph{Internet of Things}}  \bibinfo{volume}{26} (\bibinfo{year}{2024}), \bibinfo{pages}{101189}.
\newblock


\bibitem[Breiman(2001)]%
        {bib:rf}
\bibfield{author}{\bibinfo{person}{Leo Breiman}.} \bibinfo{year}{2001}\natexlab{}.
\newblock \showarticletitle{Random Forests}.
\newblock \bibinfo{journal}{\emph{Mach. Learn.}} \bibinfo{volume}{45}, \bibinfo{number}{1} (\bibinfo{date}{Oct.} \bibinfo{year}{2001}), \bibinfo{pages}{5–32}.
\newblock
\showISSN{0885-6125}
\urldef\tempurl%
\url{https://doi.org/10.1023/A:1010933404324}
\showDOI{\tempurl}


\bibitem[Ciliberto et~al\mbox{.}(2021)]%
        {ciliberto2021opportunity++}
\bibfield{author}{\bibinfo{person}{Mathias Ciliberto}, \bibinfo{person}{Vitor Fortes~Rey}, \bibinfo{person}{Alberto Calatroni}, \bibinfo{person}{Paul Lukowicz}, {and} \bibinfo{person}{Daniel Roggen}.} \bibinfo{year}{2021}\natexlab{}.
\newblock \showarticletitle{Opportunity++: A multimodal dataset for video-and wearable, object and ambient sensors-based human activity recognition}.
\newblock \bibinfo{journal}{\emph{Frontiers in Computer Science}}  \bibinfo{volume}{3} (\bibinfo{year}{2021}), \bibinfo{pages}{792065}.
\newblock


\bibitem[Dwork(2006)]%
        {bib:dp}
\bibfield{author}{\bibinfo{person}{Cynthia Dwork}.} \bibinfo{year}{2006}\natexlab{}.
\newblock \showarticletitle{Differential Privacy}. In \bibinfo{booktitle}{\emph{Automata, Languages and Programming}}, \bibfield{editor}{\bibinfo{person}{Michele Bugliesi}, \bibinfo{person}{Bart Preneel}, \bibinfo{person}{Vladimiro Sassone}, {and} \bibinfo{person}{Ingo Wegener}} (Eds.). \bibinfo{publisher}{Springer Berlin Heidelberg}, \bibinfo{address}{Berlin, Heidelberg}, \bibinfo{pages}{1--12}.
\newblock
\showISBNx{978-3-540-35908-1}


\bibitem[Dwork(2008)]%
        {bib:any-prevent}
\bibfield{author}{\bibinfo{person}{Cynthia Dwork}.} \bibinfo{year}{2008}\natexlab{}.
\newblock \showarticletitle{An Ad Omnia Approach to Defining and Achieving Private Data Analysis}. In \bibinfo{booktitle}{\emph{Privacy, Security, and Trust in KDD}}, \bibfield{editor}{\bibinfo{person}{Francesco Bonchi}, \bibinfo{person}{Elena Ferrari}, \bibinfo{person}{Bradley Malin}, {and} \bibinfo{person}{Y{\"u}cel Saygin}} (Eds.). \bibinfo{publisher}{Springer Berlin Heidelberg}, \bibinfo{address}{Berlin, Heidelberg}, \bibinfo{pages}{1--13}.
\newblock
\showISBNx{978-3-540-78478-4}


\bibitem[Fu et~al\mbox{.}(2024)]%
        {bib:lgbm}
\bibfield{author}{\bibinfo{person}{Pengxi Fu}, \bibinfo{person}{Jianxin Guo}, \bibinfo{person}{Hongxiang Luo}, {et~al\mbox{.}}} \bibinfo{year}{2024}\natexlab{}.
\newblock \showarticletitle{LightGBM for Human Activity Recognition Using Wearable Sensors}.
\newblock \bibinfo{journal}{\emph{Automation and Machine Learning}} \bibinfo{volume}{5}, \bibinfo{number}{1} (\bibinfo{year}{2024}), \bibinfo{pages}{113--118}.
\newblock


\bibitem[Fujimoto et~al\mbox{.}(2023)]%
        {bib:fujimoto}
\bibfield{author}{\bibinfo{person}{Ryusei Fujimoto}, \bibinfo{person}{Yugo Nakamura}, {and} \bibinfo{person}{Yutaka Arakawa}.} \bibinfo{year}{2023}\natexlab{}.
\newblock \showarticletitle{Differential Privacy with Weighted $\epsilon$ for Privacy-Preservation in Human Activity Recognition}. In \bibinfo{booktitle}{\emph{2023 IEEE International Conference on Pervasive Computing and Communications Workshops and other Affiliated Events (PerCom Workshops, PrivaCom 2023)}} (2023-03-13).
\newblock


\bibitem[Garain et~al\mbox{.}(2022)]%
        {bib:har}
\bibfield{author}{\bibinfo{person}{Avishek Garain}, \bibinfo{person}{Rudrajit Dawn}, \bibinfo{person}{Saswat Singh}, {and} \bibinfo{person}{Chandreyee Chowdhury}.} \bibinfo{year}{2022}\natexlab{}.
\newblock \showarticletitle{Differentially private human activity recognition for smartphone users}.
\newblock \bibinfo{journal}{\emph{Multimedia Tools and Applications}}  \bibinfo{volume}{81} (\bibinfo{date}{05} \bibinfo{year}{2022}).
\newblock
\urldef\tempurl%
\url{https://doi.org/10.1007/s11042-022-13185-4}
\showDOI{\tempurl}


\bibitem[Goodfellow et~al\mbox{.}(2016)]%
        {bib:autoencoder}
\bibfield{author}{\bibinfo{person}{Ian Goodfellow}, \bibinfo{person}{Yoshua Bengio}, {and} \bibinfo{person}{Aaron Courville}.} \bibinfo{year}{2016}\natexlab{}.
\newblock \bibinfo{booktitle}{\emph{Deep Learning}}.
\newblock \bibinfo{publisher}{MIT Press}.
\newblock
\newblock
\shownote{\url{http://www.deeplearningbook.org}}.


\bibitem[Hindistan and Yetkin(2023)]%
        {bib:hybrid-dp-gan}
\bibfield{author}{\bibinfo{person}{Yavuz~Selim Hindistan} {and} \bibinfo{person}{E.~Fatih Yetkin}.} \bibinfo{year}{2023}\natexlab{}.
\newblock \showarticletitle{A Hybrid Approach With GAN and DP for Privacy Preservation of IIoT Data}.
\newblock \bibinfo{journal}{\emph{IEEE Access}}  \bibinfo{volume}{11} (\bibinfo{year}{2023}), \bibinfo{pages}{5837--5849}.
\newblock
\urldef\tempurl%
\url{https://doi.org/10.1109/ACCESS.2023.3235969}
\showDOI{\tempurl}


\bibitem[Huang et~al\mbox{.}(2020)]%
        {bib:cnn-1}
\bibfield{author}{\bibinfo{person}{Jiahui Huang}, \bibinfo{person}{Shuisheng Lin}, \bibinfo{person}{Ning Wang}, \bibinfo{person}{Guanghai Dai}, \bibinfo{person}{Yuxiang Xie}, {and} \bibinfo{person}{Jun Zhou}.} \bibinfo{year}{2020}\natexlab{}.
\newblock \showarticletitle{TSE-CNN: A Two-Stage End-to-End CNN for Human Activity Recognition}.
\newblock \bibinfo{journal}{\emph{IEEE Journal of Biomedical and Health Informatics}} \bibinfo{volume}{24}, \bibinfo{number}{1} (\bibinfo{year}{2020}), \bibinfo{pages}{292--299}.
\newblock
\urldef\tempurl%
\url{https://doi.org/10.1109/JBHI.2019.2909688}
\showDOI{\tempurl}


\bibitem[Iman et~al\mbox{.}(2024)]%
        {bib:adpcm-2}
\bibfield{author}{\bibinfo{person}{Khalid~ismail Iman}, \bibinfo{person}{R.~I. Boby}, \bibinfo{person}{Md.~Sazib Mollik}, \bibinfo{person}{Imran Chowdhury}, \bibinfo{person}{Sahidur Rahman}, \bibinfo{person}{Wobeydul~Haq Ankur}, {and} \bibinfo{person}{Khaizuran Abdullah}.} \bibinfo{year}{2024}\natexlab{}.
\newblock \showarticletitle{A Real-Time IoT Cryptograph Communication of Chaotic-ADPCM Coding Method for IoT Applications}. In \bibinfo{booktitle}{\emph{2024 6th International Conference on Electrical Engineering and Information \& Communication Technology (ICEEICT)}}. \bibinfo{pages}{1401--1406}.
\newblock
\urldef\tempurl%
\url{https://doi.org/10.1109/ICEEICT62016.2024.10534526}
\showDOI{\tempurl}


\bibitem[Ji et~al\mbox{.}(2017)]%
        {bib:id_removal}
\bibfield{author}{\bibinfo{person}{Shouling Ji}, \bibinfo{person}{Prateek Mittal}, {and} \bibinfo{person}{Raheem Beyah}.} \bibinfo{year}{2017}\natexlab{}.
\newblock \showarticletitle{Graph Data Anonymization, De-Anonymization Attacks, and De-Anonymizability Quantification: A Survey}.
\newblock \bibinfo{journal}{\emph{IEEE Communications Surveys \& Tutorials}} \bibinfo{volume}{19}, \bibinfo{number}{2} (\bibinfo{year}{2017}), \bibinfo{pages}{1305--1326}.
\newblock
\urldef\tempurl%
\url{https://doi.org/10.1109/COMST.2016.2633620}
\showDOI{\tempurl}


\bibitem[Malekzadeh et~al\mbox{.}(2019)]%
        {bib:motion-sense}
\bibfield{author}{\bibinfo{person}{Mohammad Malekzadeh}, \bibinfo{person}{Richard~G. Clegg}, \bibinfo{person}{Andrea Cavallaro}, {and} \bibinfo{person}{Hamed Haddadi}.} \bibinfo{year}{2019}\natexlab{}.
\newblock \showarticletitle{Mobile Sensor Data Anonymization}. In \bibinfo{booktitle}{\emph{Proceedings of the International Conference on Internet of Things Design and Implementation}} (Montreal, Quebec, Canada) \emph{(\bibinfo{series}{IoTDI '19})}. \bibinfo{publisher}{ACM}, \bibinfo{address}{New York, NY, USA}, \bibinfo{pages}{49--58}.
\newblock
\showISBNx{978-1-4503-6283-2}
\urldef\tempurl%
\url{https://doi.org/10.1145/3302505.3310068}
\showDOI{\tempurl}


\bibitem[Malekzadeh et~al\mbox{.}(2020)]%
        {bib:AAE}
\bibfield{author}{\bibinfo{person}{Mohammad Malekzadeh}, \bibinfo{person}{Richard~G. Clegg}, \bibinfo{person}{Andrea Cavallaro}, {and} \bibinfo{person}{Hamed Haddadi}.} \bibinfo{year}{2020}\natexlab{}.
\newblock \showarticletitle{Privacy and utility preserving sensor-data transformations}.
\newblock \bibinfo{journal}{\emph{Pervasive and Mobile Computing}}  \bibinfo{volume}{63} (\bibinfo{year}{2020}), \bibinfo{pages}{101132}.
\newblock
\showISSN{1574-1192}
\urldef\tempurl%
\url{https://doi.org/10.1016/j.pmcj.2020.101132}
\showDOI{\tempurl}


\bibitem[Mekruksavanich and Jitpattanakul(2021)]%
        {bib:wearable_uid}
\bibfield{author}{\bibinfo{person}{Sakorn Mekruksavanich} {and} \bibinfo{person}{Anuchit Jitpattanakul}.} \bibinfo{year}{2021}\natexlab{}.
\newblock \showarticletitle{Biometric User Identification Based on Human Activity Recognition Using Wearable Sensors: An Experiment Using Deep Learning Models}.
\newblock \bibinfo{journal}{\emph{Electronics}} \bibinfo{volume}{10}, \bibinfo{number}{3} (\bibinfo{year}{2021}).
\newblock
\showISSN{2079-9292}
\urldef\tempurl%
\url{https://doi.org/10.3390/electronics10030308}
\showDOI{\tempurl}


\bibitem[Menasria et~al\mbox{.}(2022)]%
        {bib:pgan}
\bibfield{author}{\bibinfo{person}{Soumia Menasria}, \bibinfo{person}{Mingming Lu}, {and} \bibinfo{person}{Abdelghani Dahou}.} \bibinfo{year}{2022}\natexlab{}.
\newblock \showarticletitle{PGAN framework for synthesizing sensor data privately}.
\newblock \bibinfo{journal}{\emph{Journal of Information Security and Applications}}  \bibinfo{volume}{67} (\bibinfo{year}{2022}), \bibinfo{pages}{103204}.
\newblock
\showISSN{2214-2126}
\urldef\tempurl%
\url{https://doi.org/10.1016/j.jisa.2022.103204}
\showDOI{\tempurl}


\bibitem[Papernot et~al\mbox{.}(2016)]%
        {bib:pate}
\bibfield{author}{\bibinfo{person}{Nicolas Papernot}, \bibinfo{person}{Mart{\'i}n Abadi}, \bibinfo{person}{{\'U}lfar Erlingsson}, \bibinfo{person}{Ian~J. Goodfellow}, {and} \bibinfo{person}{Kunal Talwar}.} \bibinfo{year}{2016}\natexlab{}.
\newblock \showarticletitle{Semi-supervised Knowledge Transfer for Deep Learning from Private Training Data}.
\newblock \bibinfo{journal}{\emph{ArXiv}}  \bibinfo{volume}{abs/1610.05755} (\bibinfo{year}{2016}).
\newblock
\urldef\tempurl%
\url{https://api.semanticscholar.org/CorpusID:8696462}
\showURL{%
\tempurl}


\bibitem[Parra-Arnau et~al\mbox{.}(2022)]%
        {bib:adpcm}
\bibfield{author}{\bibinfo{person}{Javier Parra-Arnau}, \bibinfo{person}{Thorsten Strufe}, {and} \bibinfo{person}{Josep Domingo-Ferrer}.} \bibinfo{year}{2022}\natexlab{}.
\newblock \showarticletitle{Differentially private publication of database streams via hybrid video coding}.
\newblock \bibinfo{journal}{\emph{Knowledge-Based Systems}}  \bibinfo{volume}{247} (\bibinfo{year}{2022}), \bibinfo{pages}{108778}.
\newblock
\showISSN{0950-7051}
\urldef\tempurl%
\url{https://doi.org/10.1016/j.knosys.2022.108778}
\showDOI{\tempurl}


\bibitem[Radford et~al\mbox{.}(2015)]%
        {bib:gan}
\bibfield{author}{\bibinfo{person}{Alec Radford}, \bibinfo{person}{Luke Metz}, {and} \bibinfo{person}{Soumith Chintala}.} \bibinfo{year}{2015}\natexlab{}.
\newblock \bibinfo{title}{Unsupervised Representation Learning with Deep Convolutional Generative Adversarial Networks}.
\newblock , \bibinfo{numpages}{16}~pages.
\newblock
\urldef\tempurl%
\url{http://arxiv.org/abs/1511.06434}
\showURL{%
\tempurl}
\newblock
\shownote{cite arxiv:1511.06434Comment: Under review as a conference paper at ICLR 2016}.


\bibitem[Reiss(2012)]%
        {bib:pamap2}
\bibfield{author}{\bibinfo{person}{Attila Reiss}.} \bibinfo{year}{2012}\natexlab{}.
\newblock \bibinfo{title}{{PAMAP2 Physical Activity Monitoring}}.
\newblock \bibinfo{howpublished}{UCI Machine Learning Repository}.
\newblock
\newblock
\shownote{{DOI}: https://doi.org/10.24432/C5NW2H}.


\bibitem[Roy and Girdzijauskas(2023)]%
        {bib:har-2}
\bibfield{author}{\bibinfo{person}{D. Roy} {and} \bibinfo{person}{S. Girdzijauskas}.} \bibinfo{year}{2023}\natexlab{}.
\newblock \showarticletitle{Temporal Differential Privacy for Human Activity Recognition}.
\newblock \bibinfo{journal}{\emph{the 2023 IEEE 10th International Conference on Data Science and Advanced Analytics (DSAA)}} (\bibinfo{date}{10} \bibinfo{year}{2023}), \bibinfo{pages}{1--10}.
\newblock
\urldef\tempurl%
\url{https://doi.org/10.1109/DSAA60987}
\showDOI{\tempurl}


\bibitem[Sikder and Nahid(2021)]%
        {sikder2021ku}
\bibfield{author}{\bibinfo{person}{Niloy Sikder} {and} \bibinfo{person}{Abdullah-Al Nahid}.} \bibinfo{year}{2021}\natexlab{}.
\newblock \showarticletitle{KU-HAR: An open dataset for heterogeneous human activity recognition}.
\newblock \bibinfo{journal}{\emph{Pattern Recognition Letters}}  \bibinfo{volume}{146} (\bibinfo{year}{2021}), \bibinfo{pages}{46--54}.
\newblock


\bibitem[Steil et~al\mbox{.}(2019)]%
        {bib:eye_track}
\bibfield{author}{\bibinfo{person}{Julian Steil}, \bibinfo{person}{Inken Hagestedt}, \bibinfo{person}{Michael~Xuelin Huang}, {and} \bibinfo{person}{Andreas Bulling}.} \bibinfo{year}{2019}\natexlab{}.
\newblock \showarticletitle{Privacy-Aware Eye Tracking Using Differential Privacy}. In \bibinfo{booktitle}{\emph{Proceedings of the 11th ACM Symposium on Eye Tracking Research \& Applications}} (Denver, Colorado) \emph{(\bibinfo{series}{ETRA '19})}. \bibinfo{publisher}{Association for Computing Machinery}, \bibinfo{address}{New York, NY, USA}, Article \bibinfo{articleno}{27}, \bibinfo{numpages}{9}~pages.
\newblock
\showISBNx{9781450367097}
\urldef\tempurl%
\url{https://doi.org/10.1145/3314111.3319915}
\showURL{%
\tempurl}


\bibitem[Stirapongsasuti et~al\mbox{.}(2024)]%
        {stirapongsasuti2024preserving}
\bibfield{author}{\bibinfo{person}{Sopicha Stirapongsasuti}, \bibinfo{person}{Francis~Jerome Tiausas}, \bibinfo{person}{Yugo Nakamura}, {and} \bibinfo{person}{Keiichi Yasumoto}.} \bibinfo{year}{2024}\natexlab{}.
\newblock \showarticletitle{Preserving Data Utility in Differentially Private Smart Home Data}.
\newblock \bibinfo{journal}{\emph{IEEE Access}} (\bibinfo{year}{2024}).
\newblock


\bibitem[Sweeney(2002)]%
        {bib:k-anonymity}
\bibfield{author}{\bibinfo{person}{Latanya Sweeney}.} \bibinfo{year}{2002}\natexlab{}.
\newblock \showarticletitle{Achieving k-anonymity privacy protection using generalization and suppression}.
\newblock \bibinfo{journal}{\emph{International Journal of Uncertainty, Fuzziness and Knowledge-Based Systems}} \bibinfo{volume}{10}, \bibinfo{number}{05} (\bibinfo{year}{2002}), \bibinfo{pages}{571--588}.
\newblock


\bibitem[Tripathy et~al\mbox{.}(2019)]%
        {bib:gan2}
\bibfield{author}{\bibinfo{person}{Ardhendu Tripathy}, \bibinfo{person}{Ye Wang}, {and} \bibinfo{person}{Prakash Ishwar}.} \bibinfo{year}{2019}\natexlab{}.
\newblock \showarticletitle{Privacy-Preserving Adversarial Networks}. In \bibinfo{booktitle}{\emph{2019 57th Annual Allerton Conference on Communication, Control, and Computing (Allerton)}} (Monticello, IL, USA). \bibinfo{publisher}{IEEE Press}, \bibinfo{pages}{495–505}.
\newblock
\urldef\tempurl%
\url{https://doi.org/10.1109/ALLERTON.2019.8919758}
\showDOI{\tempurl}


\bibitem[Vavoulas et~al\mbox{.}(2016)]%
        {vavoulas2016mobiact}
\bibfield{author}{\bibinfo{person}{George Vavoulas}, \bibinfo{person}{Charikleia Chatzaki}, \bibinfo{person}{Thodoris Malliotakis}, \bibinfo{person}{Matthew Pediaditis}, {and} \bibinfo{person}{Manolis Tsiknakis}.} \bibinfo{year}{2016}\natexlab{}.
\newblock \showarticletitle{The mobiact dataset: Recognition of activities of daily living using smartphones}. In \bibinfo{booktitle}{\emph{International conference on information and communication technologies for ageing well and e-health}}, Vol.~\bibinfo{volume}{2}. SciTePress, \bibinfo{pages}{143--151}.
\newblock


\bibitem[Vecchio et~al\mbox{.}(2017)]%
        {bib:wearable_har}
\bibfield{author}{\bibinfo{person}{Alessio Vecchio}, \bibinfo{person}{Federico Mulas}, {and} \bibinfo{person}{Guglielmo Cola}.} \bibinfo{year}{2017}\natexlab{}.
\newblock \showarticletitle{Posture Recognition Using the Interdistances Between Wearable Devices}.
\newblock \bibinfo{journal}{\emph{IEEE Sensors Letters}} \bibinfo{volume}{1}, \bibinfo{number}{4} (\bibinfo{year}{2017}), \bibinfo{pages}{1--4}.
\newblock
\urldef\tempurl%
\url{https://doi.org/10.1109/LSENS.2017.2726759}
\showDOI{\tempurl}


\bibitem[Yamagiwa and Ichinomiya(2021)]%
        {bib:adpcm-1}
\bibfield{author}{\bibinfo{person}{Shinichi Yamagiwa} {and} \bibinfo{person}{Yuma Ichinomiya}.} \bibinfo{year}{2021}\natexlab{}.
\newblock \showarticletitle{Stream-Based Visually Lossless Data Compression Applying Variable Bit-Length ADPCM Encoding}.
\newblock \bibinfo{journal}{\emph{Sensors}} \bibinfo{volume}{21}, \bibinfo{number}{13} (\bibinfo{year}{2021}).
\newblock
\showISSN{1424-8220}
\urldef\tempurl%
\url{https://doi.org/10.3390/s21134602}
\showDOI{\tempurl}


\bibitem[Zaki et~al\mbox{.}(2020)]%
        {bib:lr}
\bibfield{author}{\bibinfo{person}{Zunash Zaki}, \bibinfo{person}{Muhammad~Arif Shah}, \bibinfo{person}{Karzan Wakil}, {and} \bibinfo{person}{Falak Sher}.} \bibinfo{year}{2020}\natexlab{}.
\newblock \showarticletitle{Logistic regression based human activities recognition}.
\newblock \bibinfo{journal}{\emph{J. Mech. Contin. Math. Sci}} \bibinfo{volume}{15}, \bibinfo{number}{4} (\bibinfo{year}{2020}), \bibinfo{pages}{228--246}.
\newblock


\bibitem[Zhan et~al\mbox{.}(2022)]%
        {bib:trade-off-1}
\bibfield{author}{\bibinfo{person}{Yuting Zhan}, \bibinfo{person}{Hamed Haddadi}, \bibinfo{person}{Alex Kyllo}, {and} \bibinfo{person}{Afra Mashhadi}.} \bibinfo{year}{2022}\natexlab{}.
\newblock \showarticletitle{Privacy-Aware Human Mobility Prediction via Adversarial Networks}. In \bibinfo{booktitle}{\emph{2022 2nd International Workshop on Cyber-Physical-Human System Design and Implementation (CPHS)}}. \bibinfo{pages}{7--12}.
\newblock
\urldef\tempurl%
\url{https://doi.org/10.1109/CPHS56133.2022.9804533}
\showDOI{\tempurl}


\end{thebibliography}

\end{document}